\documentclass[journal]{IEEEtran}
\usepackage{amsmath,amsfonts}
\usepackage{algorithmic}
\usepackage{array}
\usepackage[caption=false,font=normalsize,labelfont=sf,textfont=sf]{subfig}
\usepackage{textcomp}
\usepackage{stfloats}
\usepackage{url}
\usepackage{verbatim}
\usepackage{graphicx}
\usepackage{xcolor}
\usepackage{balance}

\usepackage{graphics} % for pdf, bitmapped graphics files
\usepackage{epsfig} % for postscript graphics files
\usepackage{xcolor}
\usepackage{amsmath} % assumes amsmath package installed
\usepackage{amssymb}  % assumes amsmath package installed
\usepackage{cite}

\renewcommand{\Re}{\mathbb{R}}

\usepackage{bm}

\usepackage{hyperref}
\def\BibTeX{{\rm B\kern-.05em{\sc i\kern-.025em b}\kern-.08em
    T\kern-.1667em\lower.7ex\hbox{E}\kern-.125emX}}

\begin{document}
\title{Incorporating General Contact Surfaces in the Kinematics of Tendon-Driven Rolling-Contact Joint Mechanisms}

\author{Junhyoung Ha, \IEEEmembership{Member,~IEEE}, Chaewon Kim, and Chunwoo Kim
\thanks{This work was supported by the National Research Foundation (NRF) of Korea under the grant No. 2022R1C1C1005483 funded by the Korea government (MSIT). {\it (Corresponding author: Junhyoung Ha.)}}
\thanks{Junhyoung Ha, Chaewon Kim and Chunwoo Kim are with the Center for Healthcare Robotics, Department of AI Robot, Korea Institute of Science and Technology, Seoul, South Korea (e-mail: hjhdog1@gmail.com; kimcw@kist.re.kr; cwkim@kist.re.kr).}
\thanks{This work has been submitted to the IEEE for possible publication. Copyright may be transferred without notice, after which this version may no longer be accessible.}
}

% \markboth{Journal of \LaTeX\ Class Files,~Vol.~18, No.~9, September~2020}%
% \markboth{Journal of \LaTeX\ Class Files,~Vol.~X, No.~X, August~2024}%
% {How to Use the IEEEtran \LaTeX \ Templates}

\maketitle

%%%%%%%%%%%%%%%%%%%%%%%%%%%%%%%%%%%%%%%%%%%%%%%%%%%%%%%%%%%%%%%%%%%%%%%%%%%%%%%%
\begin{abstract}
This paper presents the first kinematic modeling of tendon-driven rolling-contact joint mechanisms with general contact surfaces subject to external loads. We derived the kinematics as a set of recursive equations and developed efficient iterative algorithms to solve for both tendon force actuation and tendon displacement actuation. The configuration predictions of the kinematics were experimentally validated using a prototype mechanism. Our MATLAB implementation of the proposed kinematic is available at \href{https://github.com/hjhdog1/RollingJoint}{\it https://github.com/hjhdog1/RollingJoint}.

\end{abstract}


%========================================================================%
%                                                                        %
%                                                                        %
%                        I N T R O D U C T I O N                         %
%                                                                        %
%                                                                        %
%========================================================================%
\section{Introduction}
\IEEEPARstart{C}ontinuum robots are increasingly being used as surgical instruments for minimally invasive surgeries (MIS) due to their dexterity, flexibility, and miniaturization potential. Continuously deforming continuum robots provide enhanced safety in the patient's anatomy compared to conventional rigid-link robots. Continuum robots have been developed using different mechanisms and actuations and can be categorized into two types depending on their bending components: (i) compliant joints and (ii) rolling joints.

The compliant joints consist of elastic bodies such as springs or patterned tubes. Such mechanisms are manipulated by continuous deformation of elastic materials. Naturally, they experience complicated continuum mechanics and their kinematic modeling requires simplifications of the complex material physics. However, these simplifications lead to typical drawbacks in manipulation accuracy; the prediction of the shape and tip position is inaccurate due to unmodeled factors. In addition, the overall stiffness of the mechanism is limited by that of the elastic bodies. This is not ideal for surgical instruments because, in different phases of a single procedure, low stiffness may be desired for safety and high stiffness for large payloads~\cite{kim2017active,dong2016novel,yeshmukhametov2019novel,wockenfuss2022design,luo2018designing,kim2019continuously,eastwood2018design,chitalia2020design}.
 
In contrast, the rolling joint mechanism comprises multiple links concatenated through rolling contacts. The contact surfaces of the neighboring links are designed such that they can roll on each other. This mechanism is generally driven by tendons passing through the links, and the bending angles of the rolling joints are controlled by the tendon forces and displacements. Since the mechanism does not rely on complicated continuum mechanics, its theoretical kinematic modeling only involves interactions between rigid components. Consequently, the conventional kinematics was simply derived from a geometric relationship between the bending angles and wire lengths~\cite{webster2010design,berthet2018rolling,kim2013stiffness}.

Several studies have focused on the rolling joint mechanisms. An interventional device is required to vary its stiffness for different purposes. A variable neutral-line mechanism was developed in ~\cite{kim2013stiffness} that changes the pre-load to control the stiffness. In \cite{you2021design}, a $\Sigma$-shaped wire path was suggested instead of the conventional parallel path for higher stiffness against lateral external loads. Some studies focused on the hysteresis behavior of the mechanism. A critical factor in the hysteresis behavior is the tendon friction. Kato et al. proved this relationship and extended the forward kinematics to incorporate a tension friction propagation model~\cite{kato2016tendon}. Other studies have attempted to enhance the reliability of the kinematic models by eliminating unmodeled factors from the physical manipulators. Contact slip is a typical example. Because most kinematic models were derived under zero-slip assumption, several researchers have investigated ways to ensure zero-slip. Hong et al. added gear tooth to the rolling surfaces~\cite{hong2020design}, which helped make more stabilized movements; however, it was challenging to manufacture it in a small size. Suh et al. suggested a pulley-less rolling joint with elastic fixtures that was useful in aligning the manipulator~\cite{suh2015design}. Zhang et al. used flexure straps on each rolling joint surface to make the relative contact points coincide when the manipulator bends and arranged the bending plane in series~\cite{zhang2021design}.

In many continuum robots, constant curvature designs are used for simplified modeling and analysis~\cite{webster2010design, li2018design} and reduced design parameters~\cite{bedell2011design, bergeles2015concentric}. However, non-constant curvature designs are also employed to improve model accuracy~\cite{mahl2014variable, huang2021kinematic}, achieve specific bending shapes~\cite{ros2019design}, and enhance the robot's elastic stability~\cite{ha2014achieving, ha2016Optimizing}. In particular, the initial design of the robot is crucial in determining the robot's configuration space, as the tip position and direction of continuum robots are controlled in a coupled manner according to this design. Therefore, in practice, the robots must be designed according to the geometry of the lesion~\cite{bedell2011design, torres2012task, burgner2013telerobotic, bergeles2015concentric, hong2018development}. In the case of the rolling-contact joint mechanisms, there have been studies to achieve the desired bending shapes. Suh et al. found an optimized arrangement of yaw-pitch joints for 2-DOF isotropic motion~\cite{suh2017design}. Kim et al. estimated the bending shape by minimizing the joint moments to determine the equilibrium position of the joint~\cite{kim2020accurate}. Kwon et al. inserted different thicknesses of flat springs between joints to achieve a different bending shape of the manipulator~\cite{kwon2022hyper}. Another approach for realizing the desired bending shapes for the rolling-joint mechanisms is to design rolling surfaces with variable curvatures. This approach has been barely studied because of the modeling complexity introduced by the general rolling surface, particularly when the mechanism is subject to external loads.

\begin{figure}[t]
  % \centering
    \includegraphics[width=0.70\columnwidth]{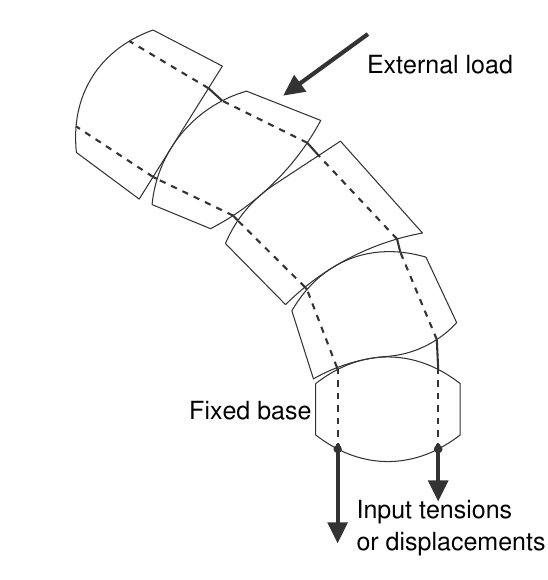}
  \caption{Tendon-driven rolling-contact joint mechanism with general contact surfaces considered in this paper. Two tendons are routed through the links, and control inputs are the tensions (i.e., pulling forces) or the displacements of the tendons.}
  \label{fig:mechanism}
\end{figure}

In this paper, we propose the first kinematic modeling of the tendon-driven rolling joint mechanism with general contact surfaces under external loads. An example is illustrated in Fig.~\ref{fig:mechanism} where the rolling surfaces are of various shapes. We formulated the kinematics as a set of recursive equations and developed an efficient Newton--Raphson algorithm to solve the equations for tendon force actuation. This algorithm was further extended into a least-squares method to account for tendon displacement actuation. The presented kinematics and algorithms were experimentally validated by comparing the configuration predictions with a physical prototype mechanism. Our MATLAB implementation of the proposed kinematics was released as an open-source repository, the link to which is provided in the Abstract.

The remainder of this paper is organized as follows. Mathematical preliminaries for kinematic formulations are presented in the following section. Subsequently, kinematic modeling is derived in Section~\ref{sec:kinematics}, followed by iterative algorithms described in Section~\ref{sec:algorithm_force} and Section~\ref{sec:algorithm_displacement}. Model validation through simulations and experiments is presented in Section~\ref{sec:sim} and Section~\ref{sec:exp}, respectively. Finally, conclusions are given in Section~\ref{sec:conclusions}.

%========================================================================%
%                                                                        %
%                       M A T H E M A T I C A L                          %
%                                                                        %
%                      P R E L I M I N A R I E S                         %
%                                                                        %
%========================================================================%
% \section{MATHEMATICAL PRELIMINARIES}
\section{Mathematical Preliminaries}

The kinematics of the rolling-contact joint mechanism involves $2$D rigid-body mechanics associated with homogeneous transformations, spatial velocities, and spatial loads. Although related formulations are found in the literature, mostly for $3$D rigid-body motions~\cite{siciliano2010robotics, lynch2017modern}, the reduction to the $2$D case is algebraically straightforward. Herein, we review a Lie-group $SE(2)$, the Euclidean group of $2$D rigid-body motions, along with its associated Lie algebra and adjoint operations.

A body-fixed frame of a rigid-body in a $2$D space is represented with a homogeneous transformation $\bm{T} \in SE(2)$:
\begin{equation}
    \bm{T} = \left[ \begin{array}{cc} \bm{R} & \bm{t} \\ \bm{0} & 1 \end{array} \right],
\end{equation}
where $\bm{R} \in SO(2)$ and $\bm{t} \in \Re^2$ are the rotation and position, respectively. Here, $SO(2)$ is the group of $2 \times 2$ rotation matrices. When the rigid-body is in rotation at an angular velocity $w \in \Re$, the angular velocity can be represented as an element of $so(2)$, the Lie algebra of $SO(2)$, as the following $2 \times 2$ skew-symmetric matrix: 
\begin{equation}
    [w] = \left[ \begin{array}{cc}
    0 & -w \\ w & 0
    \end{array} \right] \in so(2).
    \label{eqn:skew1}
\end{equation}
Given a linear velocity $\bm{v} \in \Re^2$ expressed in the body frame, the spatial velocity $\bm{\xi} = (w, \bm{v}) \in \Re^3$ is represented as an element of the Lie algebra of $SE(2)$, denoted by $se(2)$, as a $3 \times 3$ matrix below:
\begin{equation}
    [\bm{\xi}] = \left[ \begin{array}{cc} [w] & \bm{v} \\ \bm{0} & 0 \end{array} \right] \in se(2).
\end{equation}
In this case, the time derivative of $\bm{T}$ is given by
\begin{equation}
    \dot{\bm{T}} = \bm{T} [\bm{\xi}],
\end{equation}
where the upper dot represents the time derivative.
The above differentiation holds for various parameterizations.
In the context of continuum robots, an arc-length derivative is more common; in this case, $w$ is the curvature and $\bm{v}$ is the direction vector of the arc.

When a rigid-body comprises two body-fixed frames $\bm{T}_1$ and $\bm{T}_2$ with their relative transformation $\bm{T} (= \bm{T}_1^{-1} \bm{T}_2)$, the spatial velocities of $\bm{T}_1$ and $\bm{T}_2$, denoted by $\bm{\xi}_1$ and $\bm{\xi}_2$, respectively, are related to each other as follows:
\begin{equation}
    \bm{\xi}_1 = \text{Ad}_{\bm{T}} \bm{\xi}_2,
\end{equation}
or equivalently, by
\begin{equation}
    [\bm{\xi}_1] = \bm{T} [\bm{\xi}_2] \bm{T}^{-1}.
\end{equation}

Let us  consider a spatial load $\bm{\mathcal{F}}_2 = (m_2, \bm{f}_2) \in \Re^3$ expressed in $\bm{T}_2$, where $m_2 \in \Re$ is the moment and $\bm{f}_2 \in \Re^2$ is the force. The same load expressed in $\bm{T}_1$, denoted by $\bm{\mathcal{F}}_1$, is calculated using an adjoint operation $\text{Ad}^*_{\bm{T}}$ as follows:
\begin{equation}
    \bm{\mathcal{F}}_1 = \text{Ad}^*_{\bm{T}} \bm{\mathcal{F}}_2.
    \label{eqn:dAdf}
\end{equation}
Finally, consider the spatial velocity of $\bm{T}$, denoted by $\bm{\xi} = (w, \bm{v})$. The time derivative of $\text{Ad}_{\bm{T}}^*$ is given as follows:
\begin{equation}
    \dot{\text{Ad}_{\bm{T}}^*} = \text{Ad}_{\bm{T}}^* \text{ad}_{\bm{\xi}}^*.
    \label{eqn:diff_dAd}    
\end{equation}
The adjoint maps $\text{Ad}_{(\cdot)}, \text{Ad}_{(\cdot)}^*$, and $\text{ad}_{(\cdot)}^*$ are all $3 \times 3$ matrices, whose explicit forms are given in Appendix~A.

%========================================================================%
%                                                                        %
%                                                                        %
%                 K I N E M A T I C   M O D E L I N G                    %
%                                                                        %
%                                                                        %
%========================================================================%
% \section{KINEMATIC MODELING}
\section{Kinematic Modeling}
\label{sec:kinematics}

\begin{figure}[t]
  \centering
    \includegraphics[width=0.45\columnwidth]{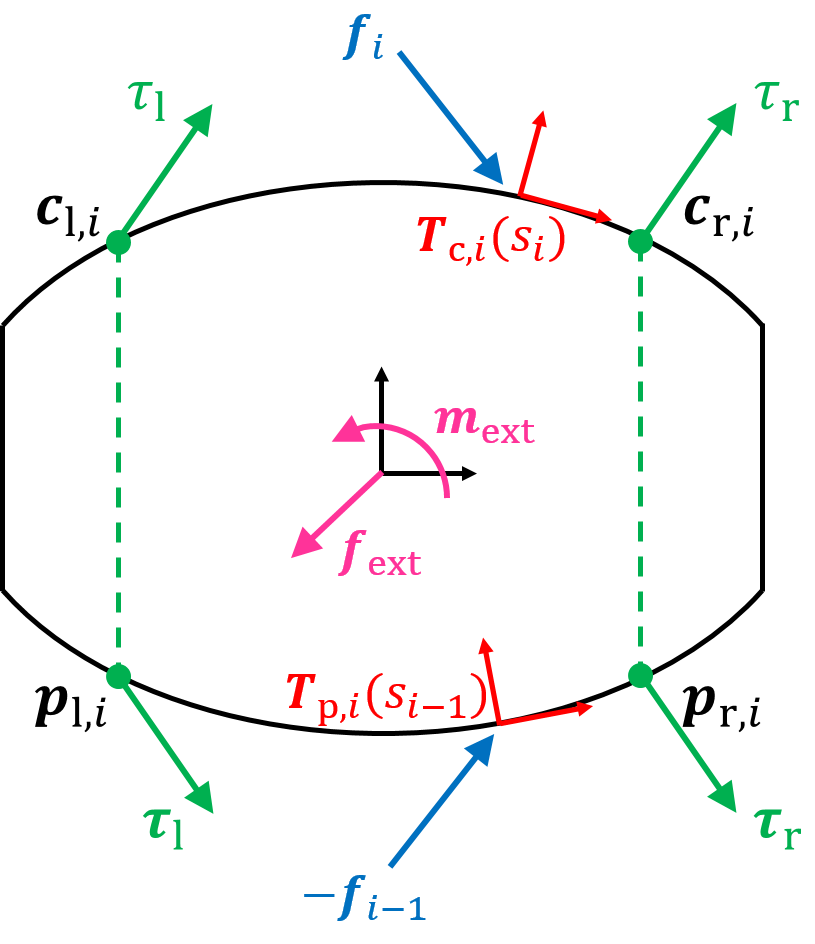}
  \caption{Variables and loads on link $i$.}
  \label{fig:link}
\end{figure}

The links in the rolling-contact joint mechanism experience tendon tensions, link-interaction forces, and external loads. The following assumptions were made for the modeling: (i) dynamic effects are ignored for quasi-static modeling, (ii) no slip is allowed between links, and (iii) friction and clearance between links and tendons are ignored.

%========================================================================%
\subsection{Link Modeling}
Link $i$ is illustrated in Fig.~\ref{fig:link}. We use the terms `{\it parent}' and `{\it child}' to refer to variables and objects associated with the lower and upper contacts, respectively. The link design parameters include
\begin{itemize}
    \item $\bm{p}_{\text{l},i}, \bm{p}_{\text{r},i}, \bm{c}_{\text{l},i}$ and $\bm{c}_{\text{r},i}$: the parent and child entry points of the left and right tendons, respectively;
    \item $\bm{T}_{\text{p}, i}(s_{i-1})$ and $\bm{T}_{\text{c} , i}(s_i)$:
    homogeneous transformations representing the parent and child contact surfaces, respectively, located on the corresponding contact surfaces with their $x$-axes tangential and $y$-axes perpendicular to the surfaces.
\end{itemize}
Here, $s_{i}$ and $s_{i-1}$ are the arc-length parameters for the child and parent surfaces, respectively. The above points and transformations are all expressed in the link's body frame (i.e., the black frame in the figure).

The loads acting on link $i$ include
\begin{itemize}
    \item $\tau_{\text{l}}$ and $\tau_{\text{r}}$: the left and right tendon tensions, respectively;
    \item $\bm{f}_{i-1}$ and $-\bm{f}_i$: the interaction forces at parent and child contacts expressed in $\bm{T}_{\text{p}, i}(s_{i-1})$ and $\bm{T}_{\text{c} , i}(s_i)$, respectively;
    \item $m_\text{ext}$ and $\bm{f}_\text{ext}$: the external moment and force, respectively, expressed in the body frame.
\end{itemize}

%========================================================================%
\subsection{Governing Equations}

Let $\bm{T}_i \in SE(2)$ denote the pose of link $i$. Given the contact point $s_{i-1}$ between link $i-1$ and link $i$, it holds
\begin{equation}
\bm{T}_i = \bm{T}_{i-1} \bm{T}_{\text{c},i-1}(s_{i-1}) \bm{T}_{\text{p},i}^{-1}(s_{i-1}).
% T_{i+1} = T_{i} T_{\text{c},i}(s_{i}) T_{\text{p},i+1}^{-1}(s_{i}).
\label{eqn:serial}
\end{equation}
Multiple loads act on link $i$, as illustrated in Fig.~\ref{fig:link}. To find a static equilibrium, we express all the loads in the body frame and sum them to zero. Based on (\ref{eqn:dAdf}), the resulting equation is given as follows:
\begin{equation}
\begin{split}
    % &\left[ \begin{array}{c} m_\text{ext,i} \\ f_\text{ext,i} \end{array} \right]
    \sum_{j \in \{ \text{l}, \text{r} \}} \left( 
      \text{Ad}_{\bm{\mathcal{I}}(\bm{c}_{j,i})}^* \left[ \begin{array}{c} 0 \\ \tau_j \hat{\bm{v}}_{j,i} \end{array} \right]
    + \text{Ad}_{\bm{\mathcal{I}}(\bm{p}_{j,i})}^* \left[ \begin{array}{c} 0 \\ \tau_j \hat{\bm{w}}_{j,i} \end{array} \right]
    \right) \\
    + \text{Ad}_{\bm{T}_{\text{p},i}}^* \left[ \begin{array}{c} 0 \\ \bm{f}_{i-1} \end{array} \right]
    - \text{Ad}_{\bm{T}_{\text{c},i}}^* \left[ \begin{array}{c} 0 \\ \bm{f}_{i} \end{array} \right]
    + \bm{\mathcal{F}}_{\text{ext},i} = 0
\end{split}
\label{eqn:equilibrium}
\end{equation}
where $\bm{\mathcal{I}}: \Re^2 \rightarrow SE(2)$ is defined as
\begin{equation}
    \bm{\mathcal{I}}(\bm{q}) = \left[ \begin{array}{cc} \bm{I} & \bm{q} \\ \bm{0} & 1 \end{array} \right],
\end{equation}
$\bm{\mathcal{F}}_{\text{ext},i} = (m_{\text{ext},i}, \bm{f}_{\text{ext},i}) \in \Re^3$ is the external load, and $\hat{\bm{v}}_{j,i} \in \Re^2$ and $\hat{\bm{w}}_{j,i} \in \Re^2$ are the unit direction vectors of tendon tensions at $\bm{c}_{j,i}$ and $\bm{p}_{j,i}$, respectively. The explicit forms of $\hat{\bm{v}}_{j,i}$ and $\hat{\bm{w}}_{j,i}$ are given as follows:
\begin{equation}
    \hat{\bm{v}}_{j,i} = \frac{\bm{v}_{j,i}}{\| \bm{v}_{j,i} \|}, \ \hat{\bm{w}}_{j,i} = \frac{\bm{w}_{j,i}}{\| \bm{w}_{j,i} \|},
\end{equation}
with $\bm{v}_{j,i}$ and $\bm{w}_{j,i}$ defined in
\begin{equation}
\begin{split}
\left[ \begin{array}{c} \bm{v}_{j,i} \\ 1 \end{array} \right] = \bm{T}_{\text{c},i} \bm{T}_{\text{p},i+1}^{-1} \left[ \begin{array}{c} \bm{p}_{j,i+1}  \\ 1 \end{array} \right] - \left[ \begin{array}{c} \bm{c}_{j,i} \\ 1 \end{array} \right], \\
\left[ \begin{array}{c} \bm{w}_{j,i} \\ 1 \end{array} \right] = \bm{T}_{\text{p},i} \bm{T}_{\text{c},i-1}^{-1} \left[ \begin{array}{c} \bm{c}_{j,i-1} \\ 1 \end{array} \right] - \left[ \begin{array}{c} \bm{p}_{j,i} \\ 1 \end{array} \right].
\label{eqn:vw}
\end{split}
\end{equation}
Note that $s_i$ and $s_{i-1}$ are implicitly present in (\ref{eqn:equilibrium}) because $\bm{T}_{\text{c},i}$ and $\hat{\bm{v}}_{j,i}$ are functions of $s_i$, and $\bm{T}_{\text{p},i}$ and $\hat{\bm{w}}_{j,i}$ are functions of $s_{i-1}$.

We remark that (\ref{eqn:serial}) and (\ref{eqn:equilibrium}) are defined for $i=2,\ldots,n$ where $n$ is the number of links. When $i=n$ in (\ref{eqn:equilibrium}), the two terms associated with the child contact, one in the summation and the other out of the summation, simply disappear because link $n$ has no child contact.

%========================================================================%
\subsection{Kinematics as a Root-Finding Problem}

Given the base link frame $\bm{T}_1$, the unknowns in the governing equations (\ref{eqn:serial}) and (\ref{eqn:equilibrium}) are $\{ \bm{T}_{i+1}, s_i, \bm{f}_i \}_{i=1,\ldots,n-1}$. Noting that an element of $SE(2)$ is $3$-DOF, there are $6(n-1)$ equations and $6(n-1)$ unknowns. Consequently, the kinematics becomes a root-finding problem with identical numbers of equations and unknowns.

%========================================================================%
%                                                                        %
%                                                                        %
%                           A L G O R I T H M                            %
%                                                                        %
%                                                                        %
%========================================================================%
% \section{ITERATIVE SOLVER FOR KINEMATICS}
\section{Iteratvie Solver of Kinematics for Tension Actuation}
\label{sec:algorithm_force}

Tendon-driven rolling-contact joint mechanisms can be actuated through either tendon forces or tendon displacements. In this section, we will first develop an iterative algorithm for force actuation and then extend it to displacement actuation in the following section.

Newton--Raphson method is a typical approach for finding roots of equations. A brute-force implementation of the Newton--Raphson method for our equations (\ref{eqn:serial}) and (\ref{eqn:equilibrium}) is directly based on the $6(n-1)$ equations and $6(n-1)$ unknowns, which involves a $6(n-1) \times 6(n-1)$ matrix inversion at every iteration. The computational load in this case exponentially increases with respect to $n$.

In this section, we present a computationally efficient version of the Newton--Raphson method by exploiting the recursiveness of the governing equations. First, we review the derivation of the Newton--Raphson method. Subsequently, we apply the same derivation to our recursive equations. 
The derived algorithm involves $n-2$ times of $3 \times 3$ matrix inversion and a single $6 \times 6$ matrix inversion at each iteration, which is much more efficient compared to the brute-force implementation.

%========================================================================%
\subsection{Review on Newton--Raphson Method}

Consider a root-finding problem for a function $h(\bm{x})$ where $h: \Re^d \rightarrow \Re^d$ and $\bm{x} \in \Re^d$. Given a perturbation $\delta \bm{x} \in \Re^d$ in $\bm{x}$, the Taylor expansion of $h(\cdot)$ is expressed as follows:
\begin{equation}
    h(\bm{x} + \delta \bm{x}) = h(\bm{x}) + \frac{\partial h}{\partial \bm{x}}(\bm{x}) \delta \bm{x} + \cdots.
    \label{eqn:Taylor}
\end{equation}
To find a root, $\delta \bm{x}$ must be found that yields $h(\bm{x} + \delta \bm{x})=0$. The Newton--Raphson method is derived by linearizing the function (i.e., discarding the second- and higher-order terms) and equating $h(\bm{x} + \delta \bm{x})$ to zero:
\begin{equation}
    \delta \bm{x} = -\frac{\partial h}{\partial \bm{x}}^{-1}(\bm{x}) h(\bm{x}).
\end{equation}
Subsequently, the update is performed by
\begin{equation}
    \bm{x} \leftarrow \bm{x} + \delta \bm{x}.
\end{equation}
For a linear function $h(\bm{x})$, the Newton--Raphson method converges at a single update, whereas it requires multiple iterations of update for nonlinear functions.

%========================================================================%
\subsection{First-Order Expansion of Governing Equations}

As reviewed previously, the Newton--Raphson method is derived based on the linearization through the first-order Taylor expansion. Similarly, we will derive the first-order Taylor expansion on our governing equations (\ref{eqn:serial}) and (\ref{eqn:equilibrium}).

Suppose that $\{s_i, \bm{f}_i\}_{i=1,\ldots,n-1}$ are given as arbitrary values and $\{ \bm{T}_i \}_{i=2,\ldots,n}$ are given accordingly by substituting $\{s_i\}$ into (\ref{eqn:serial}). In this case, the equilibrium equation (\ref{eqn:equilibrium}) does not hold. In the Newton--Raphson method, $\{s_i, \bm{f}_i\}_{i=1,\ldots,n-1}$ should be updated to compensate for the inequality of (\ref{eqn:equilibrium}).

Let us consider small perturbations in $s_i$ and $\bm{f}_i$, denoted by $\delta s_i$ and $\delta \bm{f}_i$, respectively, and the corresponding perturbations in $\bm{T}_i$, denoted by $\delta \bm{T}_i$. Further, let us define $\delta \bm{\xi}_i \in \Re^3$ as the Lie algebra associated with the perturbation, i.e., $\delta \bm{T}_i = \bm{T}_i [\delta \bm{\xi}_i]$. By substituting $\delta \bm{T}_i, \delta s_i$ and $\delta \bm{f}_i$ into (\ref{eqn:serial})-(\ref{eqn:equilibrium}) and discarding the second- and higher-order terms, the first-order Taylor expansion yields the following: 
\begin{equation}
    \delta \bm{\xi}_i = 
                    \text{Ad}_{\bm{T}_{i}^{-1} \bm{T}_{i-1}} \delta \bm{\xi}_{i-1}
                    + \text{Ad}_{\bm{T}_{\text{p},i}} 
                    (\bm{\xi}_{\text{c},i-1} - \bm{\xi}_{\text{p},i})
                    \delta s_{i-1}
\label{eqn:dserial}
\end{equation}
and
\begin{equation}
\begin{split}
    & \sum_{j \in \{ \text{l}, \text{r} \}} \left( 
    \text{Ad}_{\bm{\mathcal{I}}(\bm{c}_{j,i})}^* \left[ \begin{array}{c} 0 \\ \tau_j \frac{\partial \hat{\bm{v}}_{j,i}}{\partial s_i} \end{array} \right] \delta s_i \right. \\
    & + \left. \text{Ad}_{\bm{\mathcal{I}}(\bm{p}_{j,i})}^* \left[ \begin{array}{c} 0 \\ \tau_j \frac{\partial \hat{\bm{w}}_{j,i}}{\partial s_{i-1}} \end{array} \right] \delta s_{i-1}
    \right) \\
    & + \text{Ad}_{\bm{T}_{\text{p},i}}^* \left( \text{ad}_{\bm{\xi}_{\text{p},i}}^* \left[ \begin{array}{c} 0 \\ \bm{f}_{i-1} \end{array} \right] \delta s_{i-1}
    + \left[ \begin{array}{c} 0 \\ \delta \bm{f}_{i-1} \end{array} \right] \right) \\
    & - \text{Ad}_{\bm{T}_{\text{c},i}}^* \left( \text{ad}_{\bm{\xi}_{\text{c},i}}^* \left[ \begin{array}{c} 0 \\ \bm{f}_{i} \end{array} \right] \delta s_i
    + \left[ \begin{array}{c} 0 \\ \delta \bm{f}_{i} \end{array} \right] \right) \\
    & + \frac{\partial \bm{\mathcal{F}}_{\text{ext},i}}{\partial \bm{\xi}_i} \delta \bm{\xi}_i = -\bm{h}_i
\end{split}
\label{eqn:dequilibrium}
\end{equation}
where
\begin{equation}
\begin{split}
    \frac{\partial \hat{\bm{v}}_{j,i}}{\partial s_{i}} = \frac{I - \hat{\bm{v}}_{j,i} \hat{\bm{v}}_{j,i}^T}{\| \bm{v}_{j,i} \|}
    [u_{\text{c},i} - u_{\text{p},i+1}] \bm{R}_{i,i+1} (\bm{p}_{j,i+1} - \bm{t}_{\text{p},i+1}), \\
    \frac{\partial \hat{\bm{w}}_{j,i}}{\partial s_{i-1}} = \frac{I - \hat{\bm{w}}_{j,i} \hat{\bm{w}}_{j,i}^T}{\| \bm{w}_{j,i} \|}
    [u_{\text{p},i} - u_{\text{c},i-1}] \bm{R}_{i,i-1} (\bm{c}_{j,i-1} - \bm{t}_{\text{c},i-1}).
\end{split}
\end{equation}
Here, $\bm{\xi}_{\text{c},i-1} \in \Re^3$ and $\bm{\xi}_{\text{p},i} \in \Re^3$ are the spatial velocities of $\bm{T}_{\text{c},i-1}$ and $\bm{T}_{\text{p},i}$ along the arc-length $s_{i-1}$, respectively, given in ${\bm{T}_{\text{c},i-1}}' = \bm{T}_{\text{c},i-1} [\bm{\xi}_{\text{c},i-1}]$ and ${\bm{T}_{\text{p},i}} = \bm{T}_{\text{p},i} [\bm{\xi}_{\text{p},i}]$, where $(\cdot)'$ denotes arc-length differentiation. The variables $u_{\text{c},i-1}$ and $u_{\text{p},i}$ are the curvatures in $\bm{\xi}_{\text{c},i-1}$ and $\bm{\xi}_{\text{p},i}$, expressed as $\bm{\xi}_{\text{c},i} = (u_{\text{c},i}, \hat{\bm{e}}_x)$ and $\bm{\xi}_{\text{p},i} = (u_{\text{p},i}, \hat{\bm{e}}_x)$ where $\hat{\bm{e}}_x = [1 \ 0]^T$. The vector $\bm{h}_i \in \Re^3$ in (\ref{eqn:dequilibrium}) denotes the left side of (\ref{eqn:equilibrium}) for given $\{ \bm{T}_{i+1}, s_i, \bm{f}_i \}_{i=1,\ldots,n-1}$. Combining (\ref{eqn:dserial}) and (\ref{eqn:dequilibrium}) yields
\begin{equation}
    \left[ \begin{array}{cc}
    \bm{I} & \bm{0} \\ \bm{C}_i & \bm{D}_i
    \end{array} \right]
    \left[ \begin{array}{c}
    \delta \bm{\xi}_i \\ \delta \bm{\eta}_i
    \end{array} \right]
    + \left[ \begin{array}{cc}
    \bm{A}_i & \bm{B}_i \\ \bm{0} & \bm{E}_i
    \end{array} \right]
    \left[ \begin{array}{c}
    \delta \bm{\xi}_{i-1} \\ \delta \bm{\eta}_{i-1}
    \end{array} \right] =
    \bm{\epsilon}_i,
    \label{eqn:diff}
\end{equation}
where $\delta \bm{\eta}_i = (\delta s_i, \delta \bm{f}_i)  \in \Re^3$ and $\bm{\epsilon}_i = (\bm{0}_{3 \times 1}, -\bm{h}_i) \in \Re^6$.
The explicit forms of $\bm{A}_i, \bm{B}_i, \bm{C}_i, \bm{D}_i$ and $\bm{E}_i$ can be found in Appendix B. We remark that calculating $\bm{C}_i$ involves the differentiation of $\bm{\mathcal{F}}_{\text{ext},i}$, which is discussed in detail in Appendix C.

In (\ref{eqn:equilibrium}), the terms associated with the child contact should disappear when $i=n$. This situation is equivalently realized in (\ref{eqn:diff}) by setting $\delta \bm{\eta}_n = \bm{0}$ and $\bm{D}_n = \bm{I}$. 

%========================================================================%
\subsection{Iterative Algorithm for Tension Actuation}
\label{subsec:iterative_force}

Equation (\ref{eqn:diff}) can be expressed as follows:
\begin{equation}
    \delta \bm{x}_i = \bm{P}_i \delta \bm{x}_{i-1} + \bm{Q}_i \bm{\epsilon}_i
    \label{eqn:diff_simple}
\end{equation}
where $\delta \bm{x}_i = ( \delta \bm{\xi}_i, \delta \bm{\eta}_i) \in \Re^6$, and
\begin{equation}
\begin{split}
    \bm{Q}_i &= \left[ \begin{array}{cc}
    \bm{I} & \bm{0} \\ \bm{C}_i & \bm{D}_i
    \end{array} \right]^{-1}
    = \left[ \begin{array}{cc}
    \bm{I} & \bm{0} \\ -\bm{D}_i^{-1} \bm{C}_i & \bm{D}_i^{-1}
    \end{array} \right], \\
    \bm{P}_i &= -\bm{Q}_i
    \left[ \begin{array}{cc}
    \bm{A}_i & \bm{B}_i \\ \bm{0} & \bm{E}_i
    \end{array} \right].
\end{split}
\end{equation}
Substituting $i=2$ and $i=3$ in (\ref{eqn:diff_simple}) and combining the two equations, we get
\begin{equation}
    \delta \bm{x}_3 = \bm{P}_3 \bm{P}_2 \delta \bm{x}_1
    + \bm{P}_3 \bm{Q}_2 \bm{\epsilon}_2 + \bm{Q}_3 \bm{\epsilon}_3.
\end{equation}
Repeating this substitution recursively, the final equation is given as
\begin{equation}
    \delta \bm{x}_n = \bm{P} \delta \bm{x}_1
    + \bm{\epsilon}
\end{equation}
where
\begin{eqnarray}
    \bm{P} &=& \bm{P}_n \bm{P}_{n-1} \cdots \bm{P}_2, \\
    \bm{\epsilon} &=& \sum_{i=2}^{n-1} \bm{P}_n \cdots \bm{P}_{i+1} \bm{Q}_i \bm{\epsilon}_i
    + \bm{Q}_n \bm{\epsilon}_n.
\end{eqnarray}
By substituting $\delta \bm{\xi}_1 = \bm{0}$ and $\delta \bm{\eta}_n = \bm{0}$, we get
\begin{equation}
    \left[ \begin{array}{c}
    \delta \bm{\xi}_n \\ \delta \bm{\eta}_1
    \end{array} \right]
    = \left[ \begin{array}{cc}
        \begin{array}{c}
        \bm{I} \\ \bm{0}
        \end{array} &
        -\bm{P}_{\bm{\eta}}
    \end{array} \right]^{-1} \bm{\epsilon},
    \label{eqn:update}
\end{equation}
where $\bm{P}_{\bm{\eta}}$ is the second half of $\bm{P}$, i.e., $\bm{P} = [\bm{P}_{\bm{\xi}}, \bm{P}_{\bm{\eta}}]$.
Using $\delta \bm{\eta}_1$ acquired from (\ref{eqn:update}), all $(\delta \bm{\xi}_i, \delta \bm{\eta}_i)$ are recursively calculated by (\ref{eqn:diff_simple}). The detailed algorithm is summarized as follows:
\begin{itemize}
    \item[1)] Initialize $\{s_i, \bm{f}_i \}_{i=1,\ldots,n-1}$.
    \item[2)] Calculate $\{ \bm{T}_i \}_{i=2,\ldots,n}$ by using (\ref{eqn:serial}).
    \item[3)] Calculate $\delta \bm{\eta}_1$ and $\{ \delta \bm{\eta}_i \}_{i=2,\ldots,n-1}$ using (\ref{eqn:update}) and (\ref{eqn:diff_simple}), respectively.
    \item[4)] Update $\{s_i, \bm{f}_i \}_{i=1,\ldots,n-1}$ by $s_i \leftarrow s_i + \delta s_i$ and $\bm{f}_i \leftarrow \bm{f}_i + \delta \bm{f}_i$.
    \item[5)] Repeat 2) to 4) until (\ref{eqn:equilibrium}) holds.
\end{itemize}
Note that each iteration requires $n-2$ times of $3 \times 3$ matrix inversions for $\bm{D}_i^{-1}$ over $i = 2, \ldots, n-1$ and a $6 \times 6$ matrix inversion in (\ref{eqn:update}).

%========================================================================%
%                                                                        %
%                                                                        %
%              D I S P L A C E M E N T    A C T U A T I O N              %
%                                                                        %
%                                                                        %
%========================================================================%
\section{Iteratvie Solver of Kinematics for Displacement Actuation}
\label{sec:algorithm_displacement}

Solving the kinematics of rolling-contact joint mechanisms with tendon displacement actuation is generally challenging because the desired tendon displacements may not be achievable due to geometric constraints. For instance, pulling both tendons simultaneously is impossible since pulling one tendon requires releasing the other. As a result, obtaining an exact solution is not feasible. Instead, a least-squares approach can be employed to minimize the error between the desired and actual tendon displacements.

%========================================================================%
\subsection{Least-Squares Minimization}

Consider the tendon lengths $\bm{l} = (l_\text{l}, l_\text{r}) \in \Re^2$, representing the length of tendons in the left and right channels, respectively, from the base to the distal end:
\begin{equation}
    l_{j} = \sum_{i=1}^n \| \bm{c}_{j,i} - \bm{p}_{j,i}  \| 
                    + \sum_{i=1}^{n-1} \| \bm{v}_{j,i} \|,
    \label{eqn:tendon_length}
\end{equation}
where $j \in \{ \text{l}, \text{r} \}$, and $\bm{v}_{j,i}$ is the tendon segment between link $i$ and link $i+1$, as defined in (\ref{eqn:vw}). The actuation displacements are given as the desired tendon lengths, $\bm{l}_\text{des} = (l_{\text{l},\text{des}}, l_{\text{r},\text{des}})$.

We can formulate a least-squares problem for tendon displacement actuation, aiming to find the tendon tensions that minimize the error in tendon displacements:
\begin{equation}
    \min_{\bm{\tau}} \frac{1}{2} \| \bm{l} - \bm{l}_\text{des} \|^2,
    \label{eqn:err_min}
\end{equation}
which is subject to (\ref{eqn:serial}), (\ref{eqn:equilibrium}), and (\ref{eqn:tendon_length}), where $\bm{\tau} = (\tau_\text{l}, \tau_\text{r}) \in \Re^2$. We will solve this minimization with a gradient-based method. When updating $\bm{\tau}$, the configuration variables $\bm{T}_{i}, s_i$, and $\bm{f}_i$ must be updated accordingly. Additionally, any potential errors in (\ref{eqn:serial}) and (\ref{eqn:equilibrium}), whether from the initial solution or accumulated numerically, will be corrected by updating $\bm{T}_{i}, s_i$, and $\bm{f}_i$ as described in Section~\ref{subsec:iterative_force}.

%========================================================================%
\subsection{Tendon Lengths Jacobian with Respect to Tendon Tensions}

Next, we derive the relationship between small perturbations in tendon tensions $\bm{\tau}$ and the resulting changes in tendon lengths $\bm{l}$, expressing this relationship as a Jacobian matrix to be used in the subsequent iterative algorithm. The overall approach begins by considering a small change in $\bm{\tau}$, which induces small rollings (changes in $s_i$), eventually leading to small tendon displacements.

Let $\delta \bm{\tau}$ denote a small perturbation in $\bm{\tau}$. Similar to Section~\ref{sec:algorithm_force}, the corresponding perturbations in $(\bm{T}_{i}, s_i,\bm{f}_i)$ are denoted by $\delta \bm{x}_i$. The change in $\bm{l}$, denoted by $\delta \bm{l} = (\delta l_\text{l}, \delta l_\text{r})$, is derived through the first-order Taylor expansion of (\ref{eqn:tendon_length}). Noting that the first summation in (\ref{eqn:tendon_length}) is constant, we obtain
\begin{equation}
\delta l_j = \sum_{i=1}^{n-1} \frac{\bm{v}_{j,i}^T}{\| \bm{v}_{j,i} \|} \frac{\partial \bm{v}_{j,i}}{\partial s_{i}} \delta s_i,
\label{ref:dl}
\end{equation}
where
\begin{equation}
    \frac{\partial \bm{v}_{j,i}}{\partial s_{i}} = 
    [u_{\text{c},i} - u_{\text{p},i+1}] \bm{R}_{i,i+1} (\bm{p}_{j,i+1} - \bm{t}_{\text{p},i+1}).
\end{equation}
To relate $\delta \bm{\tau}$ and $\delta s_i$ (which is a component of $\delta \bm{x}_i$), we refer to (\ref{eqn:diff}). We assume $\bm{\epsilon}_i$ is corrected by updating $\bm{T}_{i}, s_i$, and $\bm{f}_i$ as described in Section~\ref{subsec:iterative_force}. Instead, we have a new error term on the right side of (\ref{eqn:diff}) induced by $\delta \bm{\tau}$:
\begin{equation}
    \bm{\epsilon}_i = \left[ \begin{array}{c} \bf{0}_{3 \times 2} \\
    -\bm{F}_i \end{array} \right] \delta \bm{\tau},
    \label{eqn:eps_by_tau}
\end{equation}
where
\begin{equation}
\begin{split}  
    \bm{F}_i &= [\bm{F}_{\text{l},i} \ \bm{F}_{\text{r},i}] \in \Re^{3 \times 2}, \\
    \bm{F}_{j,i} &= \text{Ad}_{\bm{\mathcal{I}}(\bm{c}_{j,i})}^* \left[ \begin{array}{c} 0 \\ \hat{\bm{v}}_{j,i} \end{array} \right]
    + \text{Ad}_{\bm{\mathcal{I}}(\bm{p}_{j,i})}^* \left[ \begin{array}{c} 0 \\ \hat{\bm{w}}_{j,i} \end{array} \right]
\end{split}
\end{equation}
for $j \in \{ \text{l}, \text{r} \}$. An impulse test approach can be employed to acquire the linear map between $\delta \bm{x}_i$ and $\delta \bm{\tau}$: $\delta \bm{x}_i$ is computed separately for $\delta \bm{\tau} = (1,0)$ and $\delta \bm{\tau} = (0,1)$ by substituting $\bm{\epsilon}_i$ from (\ref{eqn:eps_by_tau}) into (\ref{eqn:diff_simple}) and (\ref{eqn:update}), which forms the columns of the linear map $\bm{H}_i \in \Re^{6 \times 2}$, defined as
\begin{equation}
    \delta \bm{x}_i = \bm{H}_i \delta \bm{\tau}.
    \label{eqn:x_tau}
\end{equation}
Decomposing (\ref{eqn:x_tau}) only for $\delta s_i$ and substituting it into (\ref{ref:dl}) yields the Jacobian of $\bm{l}$ with respect to $\bm{\tau}$, which can be expressed as
\begin{equation}
    \delta \bm{l} = \bm{J} \delta \bm{\tau},
\end{equation}
where $\bm{J} \in \Re^{2 \times 2}$ is the Jacobian matrix.

%========================================================================%
\subsection{Iterative Algorithm for Displacement Actuation}
\label{subsec:iterative_displacement}

The gradient of the least-squares objective in (\ref{eqn:err_min}), denoted as $\bm{g}$, is given by
\begin{equation}
    \bm{g} = (\bm{l} - \bm{l}_\text{des})^T \bm{J} \in \Re^{1 \times 2}.
    \label{eqn:gradient}
\end{equation}
To update $\bm{\tau}$ in the negative gradient direction, $\delta \bm{\tau}$ can be calculated as
\begin{equation}
    \delta \bm{\tau} = - \alpha \bm{g}^T,
    \label{eqn:update_tau}
\end{equation}
where $\alpha (> 0)$ is the stepsize.

The algorithm includes a couple of additional steps compared to the algorithm in Section~\ref{sec:algorithm_force}, detailed as follows:
\begin{itemize}
    \item[1)] Initialize $\{s_i, \bm{f}_i \}_{i=1,\ldots,n-1}$.
    \item[2)] Calculate $\{ \bm{T}_i \}_{i=2,\ldots,n}$ by using (\ref{eqn:serial}).
    \item[3)] Calculate $\delta \bm{\eta}_1$ and $\{ \delta \bm{\eta}_i \}_{i=2,\ldots,n-1}$ using (\ref{eqn:update}) and (\ref{eqn:diff_simple}), respectively.
    \item[4)] Update $\{s_i, \bm{f}_i \}_{i=1,\ldots,n-1}$ by $s_i \leftarrow s_i + \delta s_i$ and $\bm{f}_i \leftarrow \bm{f}_i + \delta \bm{f}_i$.    
    \item[5)] Calculate $\bm{\tau}$ and $\{s_i, \bm{f}_i \}_{i=1,\ldots,n-1}$ using (\ref{eqn:update_tau}) and (\ref{eqn:x_tau}).
    \item[6)] Update $\bm{\tau}$ and $\{s_i, \bm{f}_i \}_{i=1,\ldots,n-1}$ by $\bm{\tau} \leftarrow \bm{\tau} + \delta \bm{\tau}$, $s_i \leftarrow s_i + \delta s_i$ and $\bm{f}_i \leftarrow \bm{f}_i + \delta \bm{f}_i$.
    \item[7)] Repeat 2) to 6) until (\ref{eqn:equilibrium}) holds and $\bm{g}$ in (\ref{eqn:gradient}) is close to $\bm{0}$.
\end{itemize}
Steps 5) and 6) are the additional steps required to minimize the least-squares error. The termination condition also includes verifying that the gradient magnitude has sufficiently decreased, ensuring convergence of the solution.
%========================================================================%
%                                                                        %
%                                                                        %
%                          S I M U L A T I O N                           %
%                                                                        %
%                                                                        %
%========================================================================%
\section{SIMULATIONS}
\label{sec:sim}

We conducted a set of simulations in the unloaded and loaded cases to verify that the model-predicted behavior of the rolling-contact joint mechanism aligns with our physical intuition.

The design of the mechanism used in the simulations is as depicted in Fig.~\ref{fig:mechanism}, the specific design parameters of which can be found in the code repository.

%========================================================================%
\subsection{Unloaded Simulations}

Intuitively, the mechanism is expected to bend toward the tendon with larger tension. In addition, because our model does not include friction, the configuration is expected to depend only on the ratio between the two tensions when there is no external load.

\begin{figure}[t]
  \subfloat[]{
    \includegraphics[width=0.41\columnwidth]{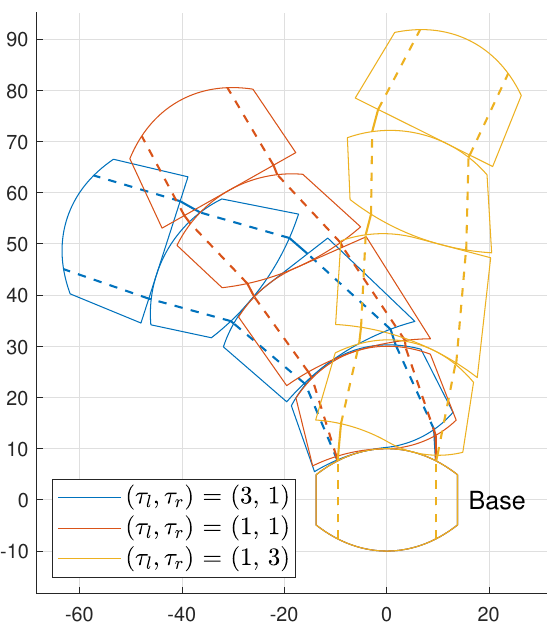}
  }
  \subfloat[]{
    \includegraphics[width=0.41\columnwidth]{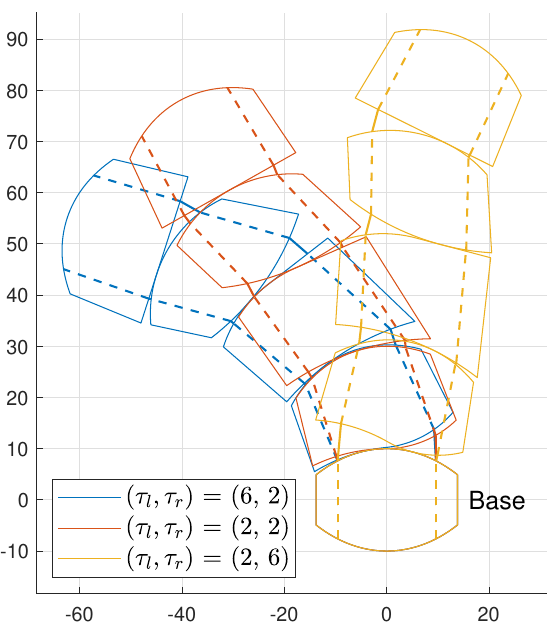}
  }
  \caption{Configuration predictions of a $5$-link mechanism without external loads. The tensions applied in (a) were $(\tau_\text{l}, \tau_\text{r}) \in \{ (3,1), (1,1), (1,3) \}$. In (b), tensions were doubled from (a), while the configurations remained identical. The force unit is arbitrary but identical between $\tau_\text{l}$ and $\tau_\text{r}$.}
  \label{fig:unloaded_sim}
\end{figure}

Fig.~\ref{fig:unloaded_sim} shows the configurations of the mechanism for various tendon tensions. As the right tension increases relative to  the left tension, the configuration leans to the right. In addition, when both tensions were doubled, the configurations remained identical to the configurations of the original tensions. These observations agree with our intuition.

%========================================================================%
\subsection{Loaded Simulations}

The configuration of the mechanism can vary from the unloaded configuration when external loads are exerted. We considered a scenario in which the same $5$-link mechanism was subjected to a pulling force to the right.

\begin{figure}[t]
  \centering
    \includegraphics[width=0.58\columnwidth]{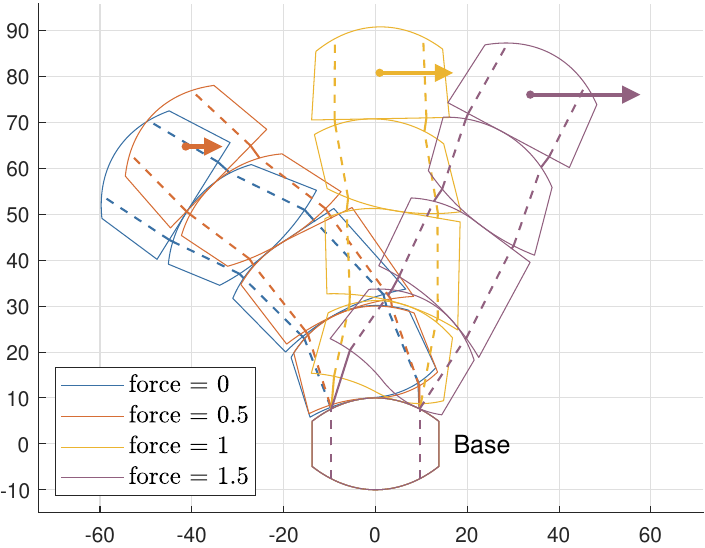}
  \caption{Configuration predictions of a $5$-link mechanism subject to an external pulling force. The tendon tensions were given as $(\tau_\text{l}, \tau_\text{r}) = (6,3)$, and the pulling force varied from $0$ to $1.5$. The forces are indicated by arrows with their lengths proportional to the force magnitudes. The force unit is arbitrary but identical to that of the tendon tensions.}
  \label{fig:loaded_sim}
\end{figure}

We kept the tendon tensions as $(\tau_\text{l}, \tau_\text{r}) = (6,3)$ and gradually increased the pulling force from $0$ to $1.5$. Again, the force unit is arbitrary but should be identical to that of the tendon forces. The simulation results are given in Fig.~\ref{fig:loaded_sim}, where the configuration gradually bent to the right as the pulling force increased.

%========================================================================%
%                                                                        %
%                                                                        %
%                           EXPERIMENTS                            %
%                                                                        %
%                                                                        %
%========================================================================%
\section{PHYSICAL EXPERIMENTS}
\label{sec:exp}

\begin{figure}[t]
    \includegraphics[width=0.39\columnwidth]{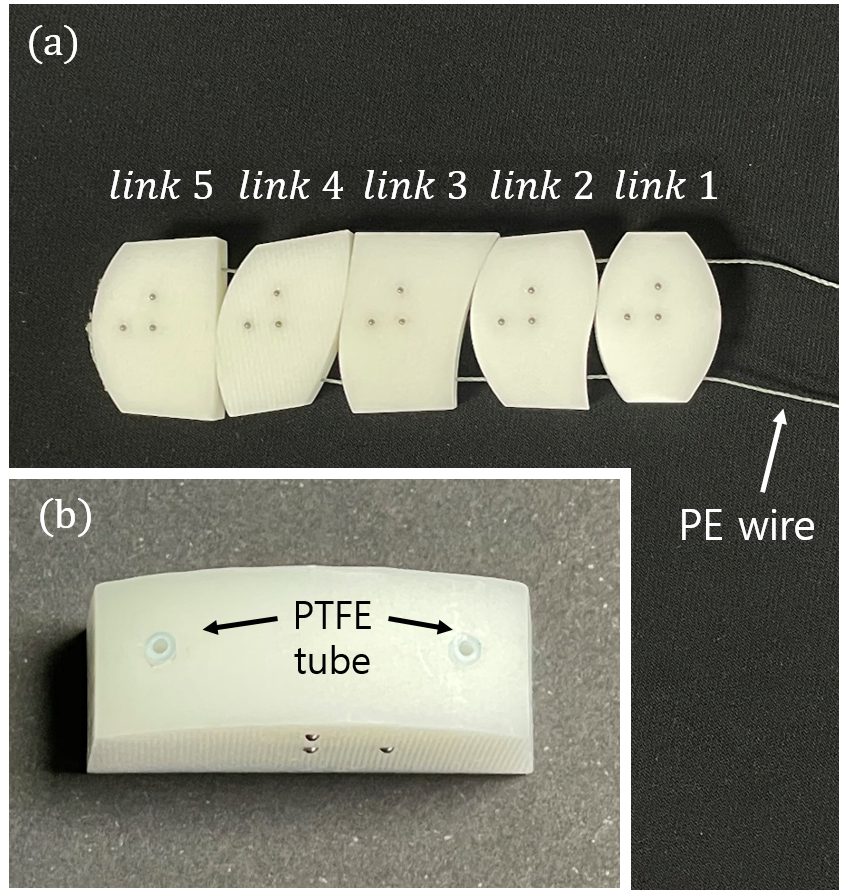}
    \includegraphics[width=0.585\columnwidth]{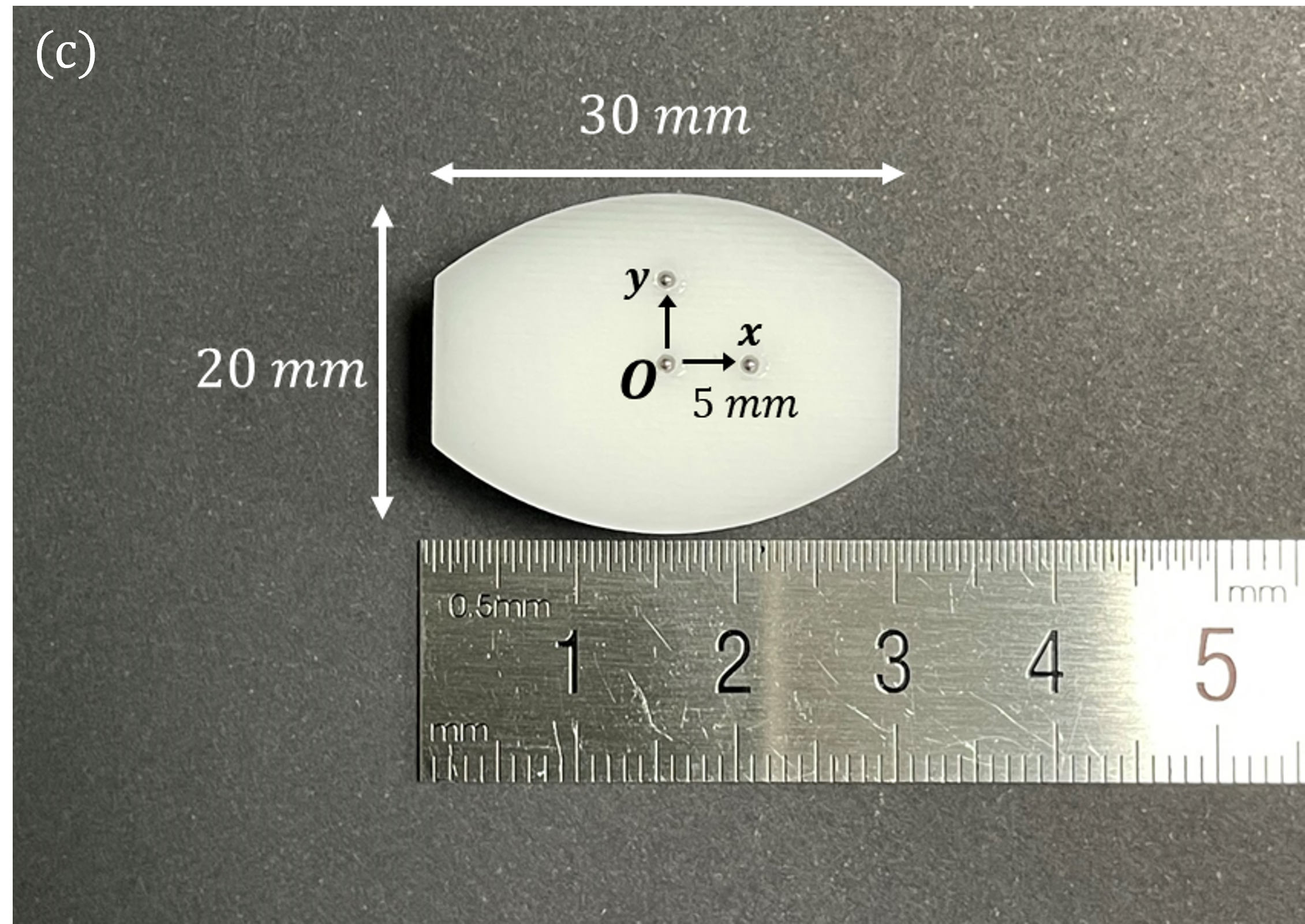}
  \caption{Prototype mechanism used in experiments: (a) manufactured mechanism, (b) vertical view of link $5$, and (c) front view of link $1$.}
  \label{fig:design}
\end{figure}

A set of physical experiments was conducted using a prototype mechanism to verify the proposed kinematic model and simulation results. We built the prototype mechanism with five links, as shown in Fig.~\ref{fig:design}(a), which has the same design as the one used in the simulations. Each link was approximately $20$ mm in length and $30$ mm in width (see Fig.~\ref{fig:design}(c)). Further, a polyethylene braided wire with a diameter of $0.28$ mm was used as the tendons, which were inserted through the PTFE  channels mounted in the links to reduce friction (see Fig.~\ref{fig:design}(b)).

In our experiments, we placed the prototype mechanism on a horizontal plane that was covered with a PTFE film to minimize the contact friction. The tendon tensions and external loads were applied by vertical weights using pulleys. To capture the configurations of the mechanism, $3$ ball markers were attached on the front face of each link, as shown in Fig.~\ref{fig:design}(c). We used the MATLAB  built-in function of the binary robust invariant scalable keypoint (BRISK) algorithm to detect the markers and calculated the pose of each link based on the marker positions.

%========================================================================%
\subsection{Unloaded Experiments}

The experiments were performed in two cases, categorized as unloaded case and loaded case. In the unloaded case, two tendons were pulled by vertical weights through pulleys without applying any external loads. The experiments aimed to compare the mechanism’s configurations with the model predictions. We varied the weight applied to each tendon from $300$ g ($2.94$ N) to $900$ g ($8.83$ N) and collected $5$ different configurations. We observed that the resulting configurations differed slightly for different initial configurations, even though the same weights were applied. This could be explained as an effect of hysteresis induced by unmodeled factors such as tendon friction. To standardize our experimental results, we collected each configuration by changing the configuration from the initial configuration with $\tau_\text{l} = 300$ g and $\tau_\text{r} = 300$ g.

We applied our model to predict the configurations for both tension actuation and displacement actuation, using the algorithms described in Section~\ref{subsec:iterative_force} and Section~\ref{subsec:iterative_displacement}, respectively. The tension values were provided by the applied weights, while the tendon displacements were obtained by measuring the lengths of tendon paths from images of each configuration.

\begin{figure}[t]
    \centering
    \subfloat[]{
        \includegraphics[width=0.085\textwidth]{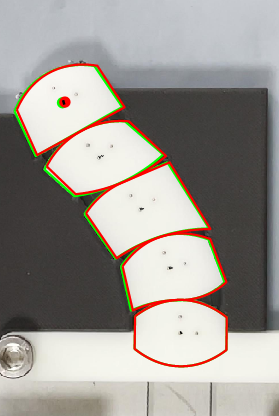}
        \label{fig:Pose1_tension}
    }
    \subfloat[]{
        \includegraphics[width=0.085\textwidth]{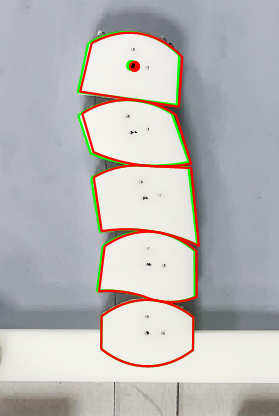}
        \label{fig:Pose2_tension}
    }
    \subfloat[]{
        \includegraphics[width=0.085\textwidth]{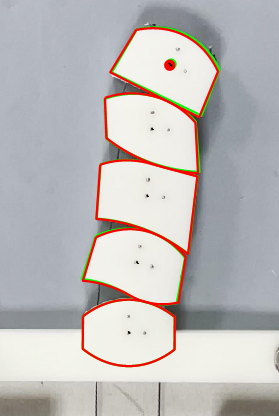}
        \label{fig:Pose3_tension}
    }
    \subfloat[]{
        \includegraphics[width=0.085\textwidth]{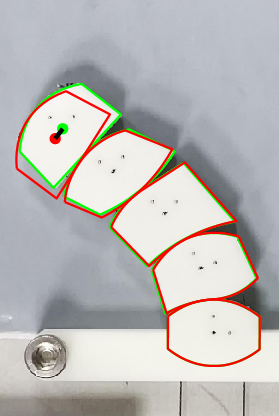}
        \label{fig:Pose4_tension}
    }
    \subfloat[]{
        \includegraphics[width=0.085\textwidth]{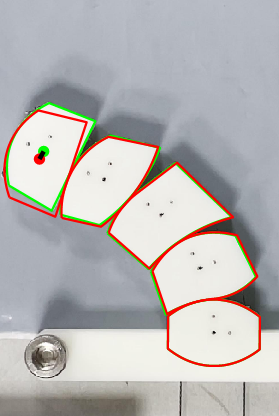}
        \label{fig:Pose5_tension}
    }
    \caption{Experimental configurations (green) without external loads compared with model predictions for tension actuation (red). The tendon tensions were given as (a) $\tau_\text{l}:\tau_\text{r}=300 \text{ g}:300 \text{ g}$ (Pose $1$, initial configuration), (b) $\tau_\text{l}:\tau_\text{r}=300 \text{ g}:600 \text{ g}$ (Pose $2$), (c) $\tau_\text{l}:\tau_\text{r}=300 \text{ g}:900 \text{ g}$ (Pose $3$), (d) $\tau_\text{l}:\tau_\text{r}=600 \text{ g}:300 \text{ g}$ (Pose $4$), and (e) $\tau_\text{l}:\tau_\text{r}=900 \text{ g}:300 \text{ g}$ (Pose $5$).}
    \label{fig:unloaded_case_tension}
\end{figure}

\begin{figure}[t]
    \centering
    \subfloat[]{
        \includegraphics[width=0.44\columnwidth]{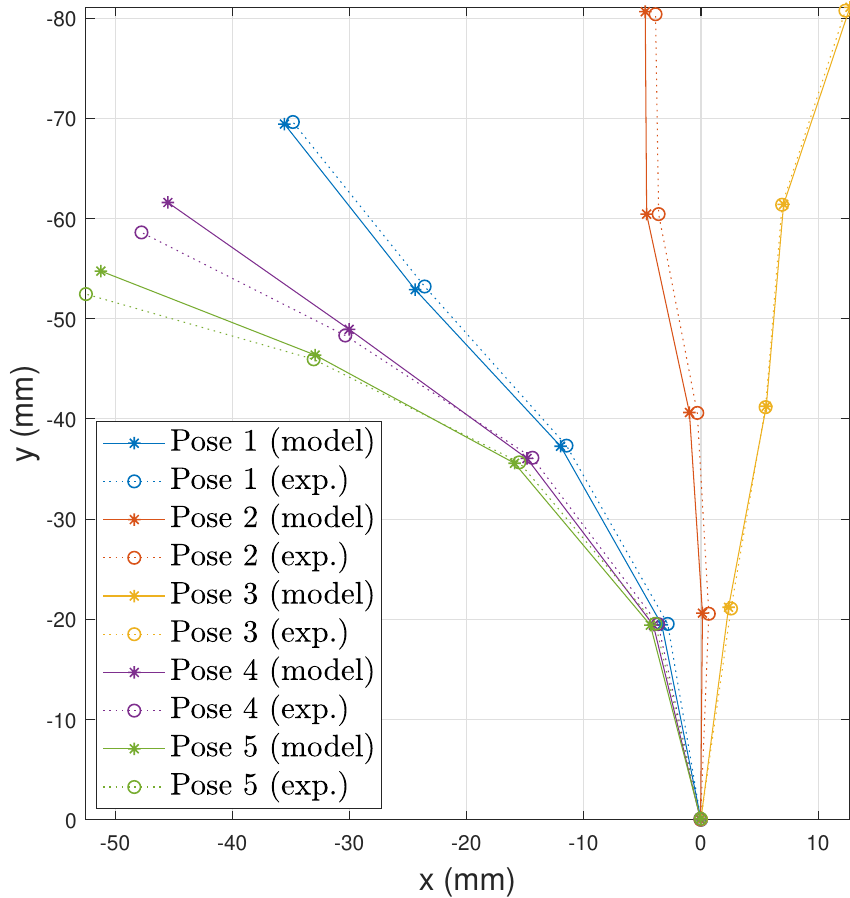}
    }
    \subfloat[]{
        \includegraphics[width=0.5\columnwidth]{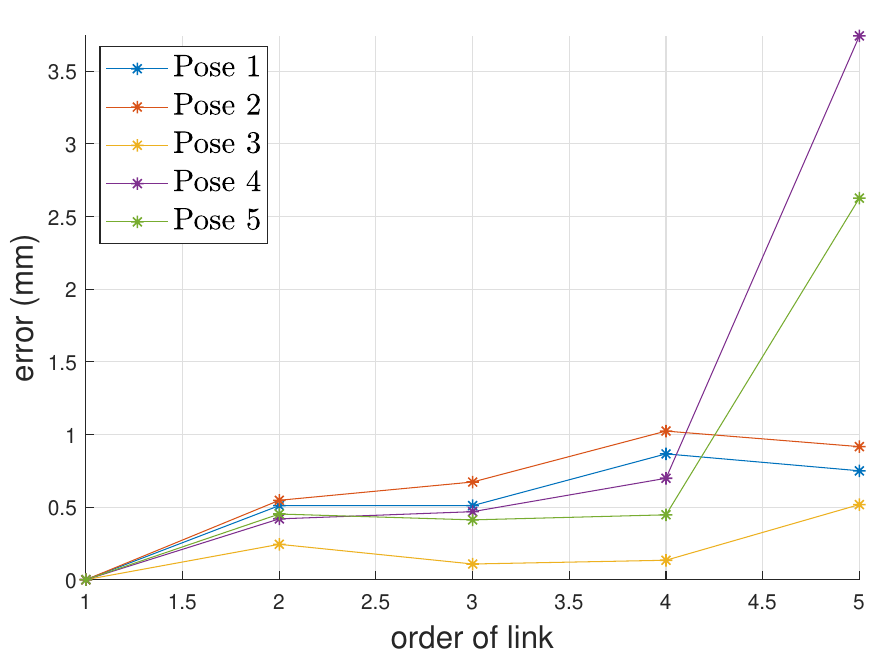}
    }
 \caption{Comparison between experimental configurations and model predictions for tension actuation in unloaded experiments: (a) link center positions of the experimental configurations and model predictions, and (b) link center position errors between the experimental configurations and model predictions. In (a), `(model)' and `(exp.)' in the legends indicate the model prediction and experimental configuration, respectively.}
 \label{fig:err_unloaded_case_tension}
\end{figure}

Fig.~\ref{fig:unloaded_case_tension} shows the resulting configurations, where the configurations predicted from the tensions were superimposed as red curves. The link center position errors are analyzed in Fig.~\ref{fig:err_unloaded_case_tension}. The link center positions are compared between the model predictions and experimental results in Fig.~\ref{fig:err_unloaded_case_tension}(a), and the link position errors for all links and configurations are shown in Fig.~\ref{fig:err_unloaded_case_tension}(b). The maximum link position error was observed to be $3.74$ mm at the configuration in Fig.~\ref{fig:unloaded_case_tension}(a), which is denoted by Pose $4$ in Fig.~\ref{fig:err_unloaded_case_tension}. The average position error of the last link across the five configurations was $1.71$ mm.

\begin{figure}[t]
    \centering
    \subfloat[]{
        \includegraphics[width=0.085\textwidth]{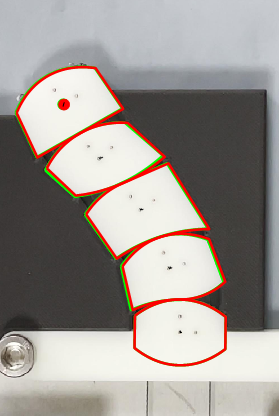}
        \label{fig:Pose1_displacement}
    }
    \subfloat[]{
        \includegraphics[width=0.085\textwidth]{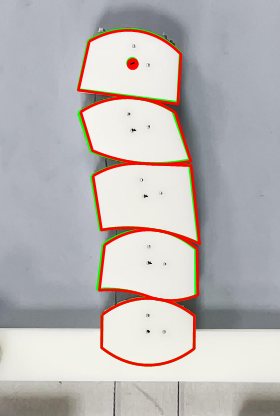}
        \label{fig:Pose2_displacement}
    }
    \subfloat[]{
        \includegraphics[width=0.085\textwidth]{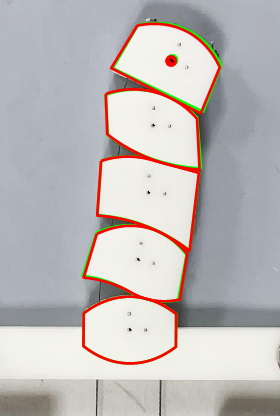}
        \label{fig:Pose3_displacement}
    }
    \subfloat[]{
        \includegraphics[width=0.085\textwidth]{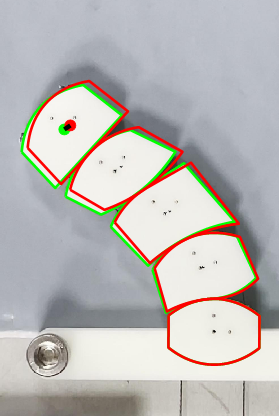}
        \label{fig:Pose4_displacement}
    }
    \subfloat[]{
        \includegraphics[width=0.085\textwidth]{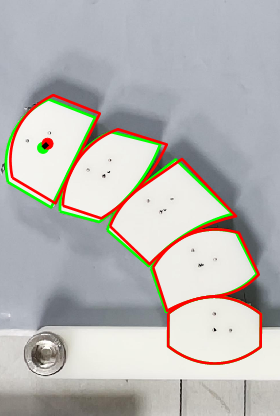}
        \label{fig:Pose5_displacement}
    }
    \caption{Experimental configurations (green) without external loads compared with model predictions for displacement actuation (red). The tendon displacements were
    measured by tracing the tendon paths in the images and are as follows: (a) $(l_\text{l}, l_\text{r}) = (90.52 \text{ mm}, 100.88 \text{ mm})$ (Pose $1$, initial configuration), (b) $(l_\text{l}, l_\text{r}) = (98.06 \text{ mm}, 95.92 \text{ mm})$ (Pose $2$), (c) $(l_\text{l}, l_\text{r}) = (102.65 \text{ mm}, 94.34 \text{ mm})$ (Pose $3$), (d) $(l_\text{l}, l_\text{r}) = (87.94 \text{ mm}, 104.17 \text{ mm})$ (Pose $4$), and (e) $(l_\text{l}, l_\text{r}) = (86.02 \text{ mm}, 107.82 \text{ mm})$ (Pose $5$).
    }
    \label{fig:unloaded_case_displacement}
\end{figure}

\begin{figure}[t]
    \centering
    \subfloat[]{
        \includegraphics[width=0.44\columnwidth]{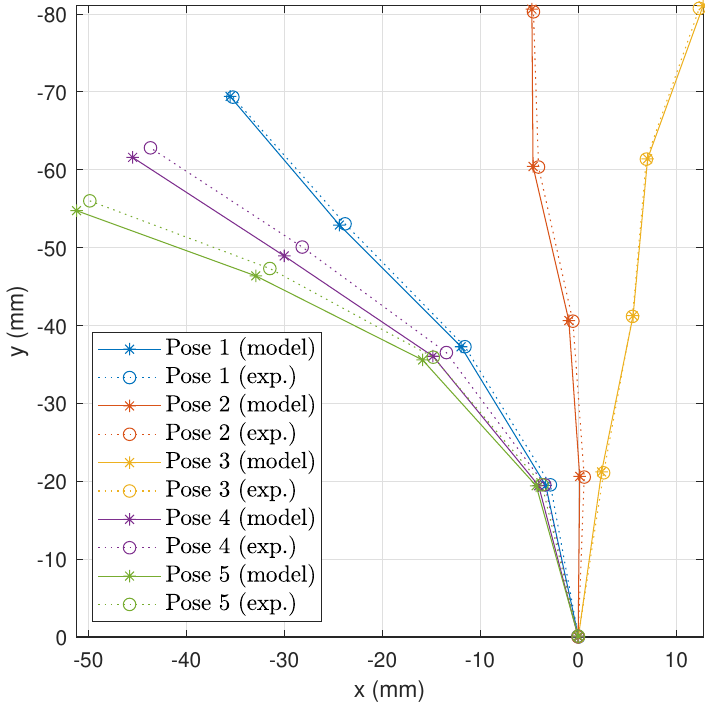}
    }
    \subfloat[]{
        \includegraphics[width=0.5\columnwidth]{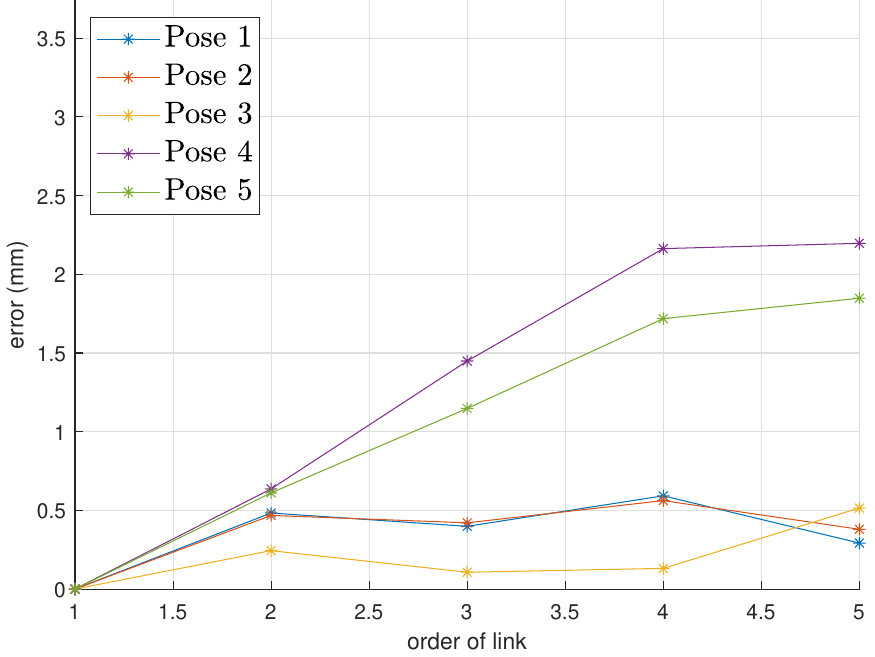}
    }
 \caption{Comparison between experimental configurations and model predictions for displacement actuation in unloaded experiments: (a) link center positions of the experimental configurations and model predictions, and (b) link center position errors between the experimental configurations and model predictions. In (a), `(model)' and `(exp.)' in the legends indicate the model prediction and experimental configuration, respectively.}
 \label{fig:err_unloaded_case_displacement}
\end{figure}

Fig.~\ref{fig:unloaded_case_displacement} and Fig.~\ref{fig:err_unloaded_case_displacement} present the results for displacement actuation. Similarly, the configurations predicted from the tendon displacements are shown in red, while the measured configurations are in green. The maximum link position error was $2.19$ mm at the configuration in Pose $4$, and the average position error of the last link across the five configurations was $1.05$ mm. We remark that the errors observed in the displacement-actuated predictions were smaller than those in the tension-actuated predictions. The larger errors in the tension-actuated cases are primarily attributed to the unmodeled tendon friction, which opposes tendon displacements and results in reduced tendon displacements compared to model predictions. In contrast, displacement-actuated cases directly utilize the tendon displacements, resulting in less uncertainty in configuration prediction.

As each configuration was collected by moving from the initial configuration with $\tau_\text{l} = 300$ g and $\tau_\text{r} = 300$ g, biases may be introduced in configurations toward the initial configuration, as can be seen in the experimental configurations in Fig.~\ref{fig:err_unloaded_case_tension}(d) and Fig.~\ref{fig:err_unloaded_case_tension}(e). We presume that the tendon with larger tension experienced large tension loss by the friction, and thus, the last link did not rotate as much as was expected for the given tensions without tension loss.

%========================================================================%
\subsection{Loaded Experiments}

Additional experiments were conducted to verify the proposed model in the presence of an external load. With tensions applied to the tendons fixed as $\tau_\text{l} = 600$ g and $\tau_\text{r} = 300$ g, an additional weight was applied as the external load to the last link through a string. The weight was incremented by $50$ g from $50$ g to $150$ g. As in the previous experiments, each configuration was collected by applying $\tau_\text{l} = 300$ g and $\tau_\text{r} = 300$ g first and substantially increasing $\tau_\text{l}$ to $\tau_\text{l} = 600$ g. The external load was applied finally.

\begin{figure}[t]
    \subfloat[]{
        \includegraphics[width=0.31\columnwidth]{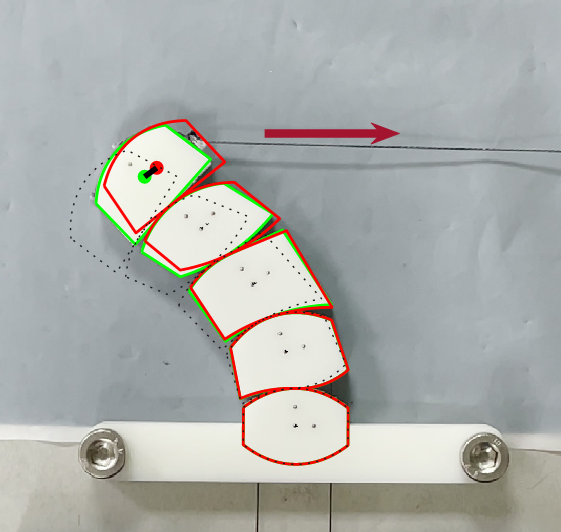}
    }
    \subfloat[]{
        \includegraphics[width=0.31\columnwidth]{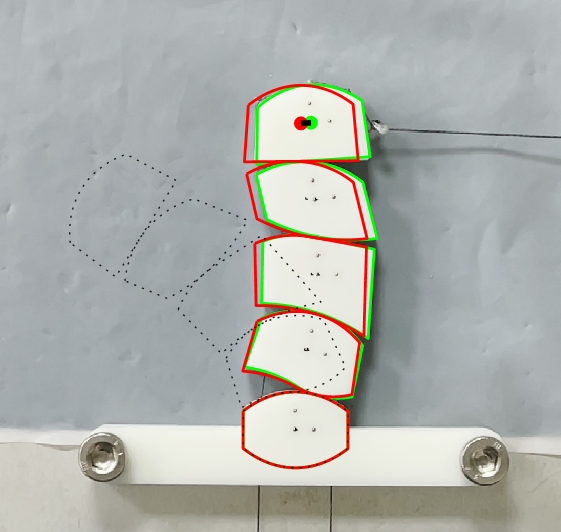}
    }
    \subfloat[]{
        \includegraphics[width=0.31\columnwidth]{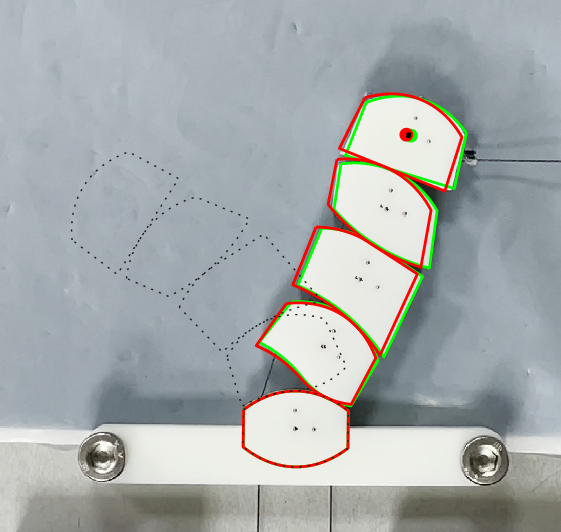}
    }
    \caption{
    Experimental configurations (green) compared to model predictions for tension actuation (red) in the presence of an external load. The black dashed curves represent the experimental configuration without the external load. The direction of the external load is indicated by a red arrow in (a). The applied loads were (a) $50$ g, (b) $100$ g, and (c) $150$ g.}
    \label{fig:loaded_case_tension}
\end{figure}

\begin{figure}[ht!]
    \subfloat[]{
        \includegraphics[width=0.44\columnwidth]{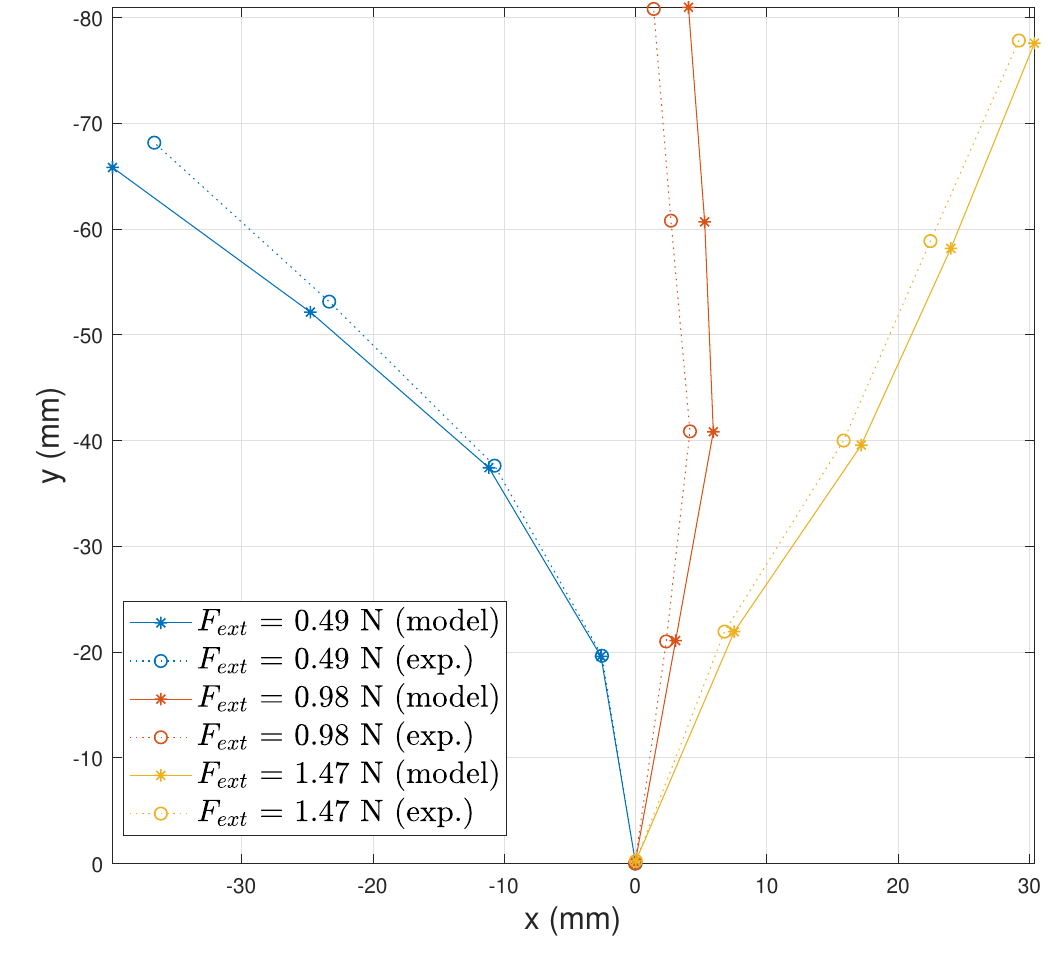}
    }    
    \subfloat[]{
        \includegraphics[width=0.5\columnwidth]{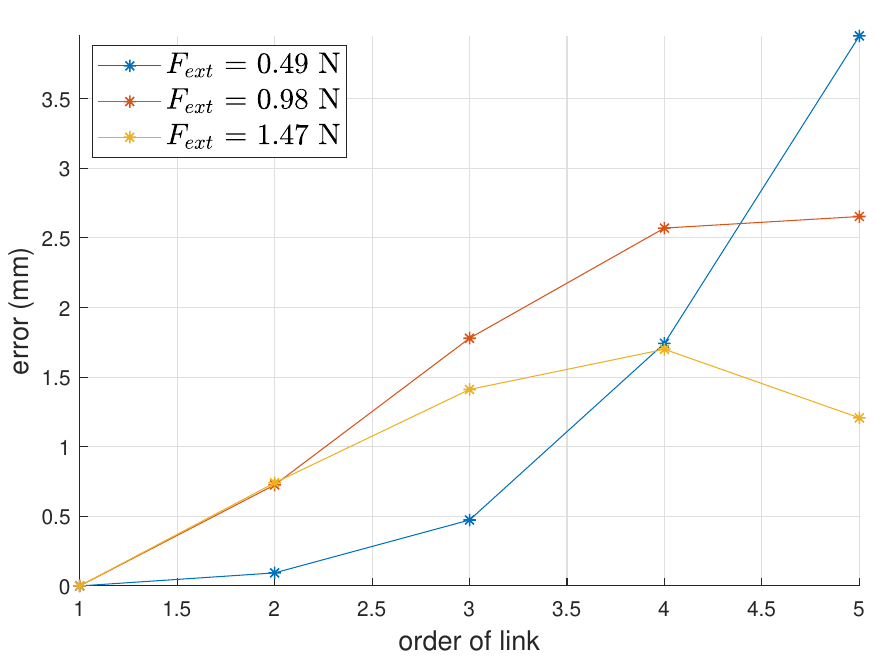}
    }
 \caption{Comparison between experimental configurations and model predictions for tension actuation in loaded experiments: (a) link center positions of the experimental configurations and model predictions, and (b) link center position errors between the experimental configurations and model predictions. The magnitude of the external load is denoted by $F_\text{ext}$ in the legends. In (a), `(model)' and `(exp.)' in the legends indicate the model prediction and experimental configuration, respectively.}
 \label{fig:err_loaded_case_tension}
\end{figure}

The collected configurations are shown in Fig.~\ref{fig:loaded_case_tension}, where the red curves represent the configurations predicted from the tensions, given the external load, and the black dashed curves indicate the unloaded configuration predicted without the external load. As observed in Fig.~\ref{fig:err_loaded_case_tension}, the maximum position error of the last link was $3.95$ mm at the configuration with the external load of $50$ g. This could be because the unloaded configuration (i.e., Pose $4$ in the unloaded experiments) already experienced a large position error. The average position error of the last link for the three configurations was $2.60$ mm.

\begin{figure}[t]
    \subfloat[]{
        \includegraphics[width=0.31\columnwidth]{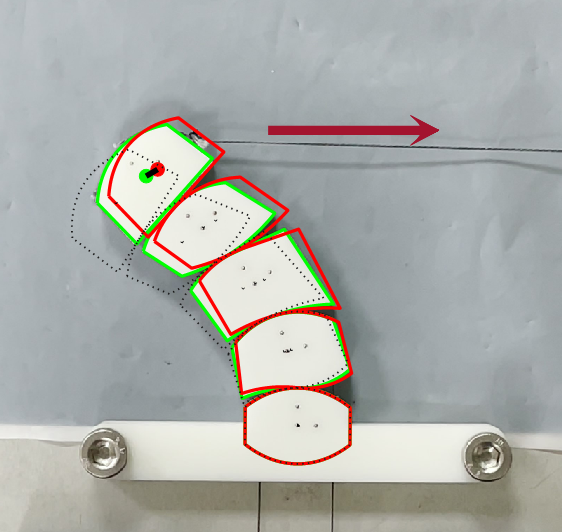}
    }
    \subfloat[]{
        \includegraphics[width=0.31\columnwidth]{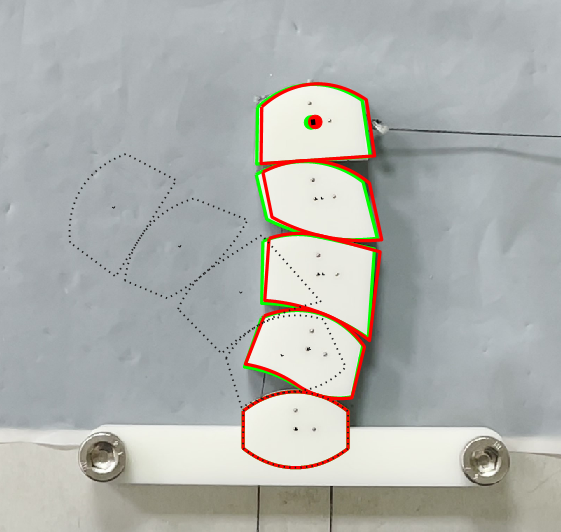}
    }
    \subfloat[]{
        \includegraphics[width=0.31\columnwidth]{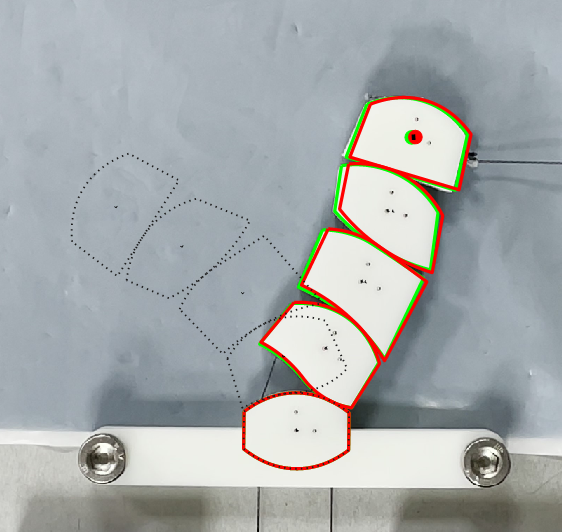}
    }
    \caption{Experimental configurations (green) compared to model predictions for displacement actuation (red) in the presence of an external load. The black dashed curves represent the experimental configuration without the external load. The direction of the external load is indicated by a red arrow in (a). The applied loads were (a) $50$ g, (b) $100$ g, and (c) $150$ g.}
    \label{fig:loaded_case_displacement}
\end{figure}

\begin{figure}[ht!]
    \subfloat[]{
        \includegraphics[width=0.44\columnwidth]{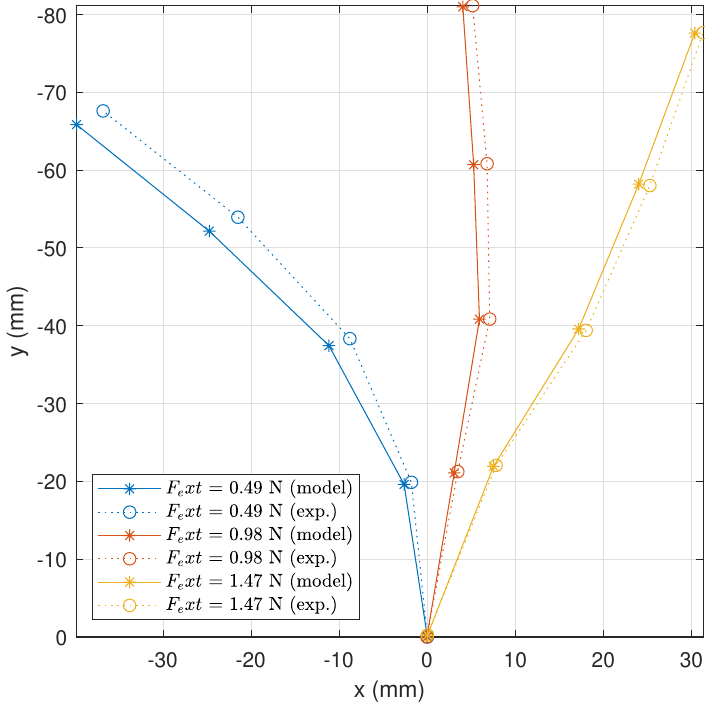}
    }    
    \subfloat[]{
        \includegraphics[width=0.5\columnwidth]{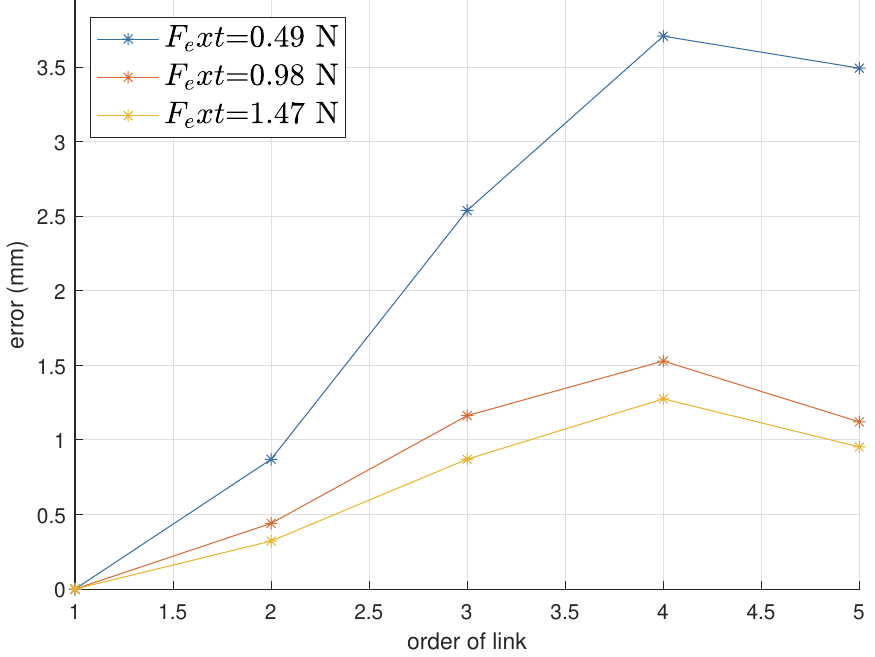}
    }
 \caption{Comparison between experimental configurations and model predictions for displacement actuation in loaded experiments: (a) link center positions of the experimental configurations and model predictions, and (b) link center position errors between the experimental configurations and model predictions. The magnitude of the external load is denoted by $F_\text{ext}$ in the legends. In (a), `(model)' and `(exp.)' in the legends indicate the model prediction and experimental configuration, respectively.}
 \label{fig:err_loaded_case_displacement}
\end{figure}

Fig.~\ref{fig:loaded_case_displacement} and Fig.~\ref{fig:err_loaded_case_displacement} present the results for displacement actuation. Similarly, the red curves in Fig.~\ref{fig:loaded_case_displacement} indicate the configurations predicted from the tendon displacements, while the green shapes represent the measured configurations. As shown in Fig.~\ref{fig:err_loaded_case_displacement}, the maximum position error of the last link was $3.49$ mm at the configuration with the external load of $50$ g, and the average position error of the last link across the three configurations was $1.86$ mm. Note that larger position errors were observed from the second last link, with a maximum error of $3.71$ mm and an average error of $2.17$ mm. Again, these errors are smaller than those observed in the tension-actuated predictions.

%========================================================================%
%                                                                        %
%                                                                        %
%                         C O N C L U S I O N S                          %
%                                                                        %
%                                                                        %
%========================================================================%
\section{CONCLUSIONS}
\label{sec:conclusions}
This paper presents the first kinematic model for tendon-driven rolling-contact joint mechanisms with general contact surfaces, which considers arbitrary external loads at arbitrary links. Two algorithms were developed to solve the model for two different types of tendon actuation: tension actuation and displacement actuation. We conducted a set of numerical and physical experiments to validate the configurations predicted by the proposed model. In our experiments with a $90$ mm-long, $5$-link mechanism, the model's prediction error for tension actuation in the center position of the last link was $3.74$ mm at most and $1.71$ mm on average in the unloaded cases and was $3.95$ mm at most and $2.60$ mm on average in the loaded cases. For displacement-actuated predictions, the center position error of the last link was $2.19$ mm at most and $1.05$ mm on average in the unloaded cases and was $3.49$ mm at most and $1.86$ mm on average in the loaded cases. We provide a MATLAB implementation of our model, which is available through the link provided in the Abstract.

%%%%%%%%%%%%%%%%%%%%%%%%%%%%%%%%%%%%%%%%%%%%%%%%%%%%%%%%%%%%%%%%%%%%%%%%%%%%%%%%
\section*{APPENDIX}

%========================================================================%
\subsection{Adjoint Operators}
\label{app:adjoint}

Consider an element of $SE(2)$ and an element of the associated Lie algebra $se(2)$:
\begin{eqnarray}
    \bm{T} &=& \left[ \begin{array}{cc} \bm{R} & \bm{t} \\ \bm{0} & 1 \end{array} \right] \in SE(2), \\
    \bm{\xi} &=& (w, \bm{v}) \in se(2).
\end{eqnarray}
In this case, the adjoint operations are given as the following $3 \times 3$ matrices:
\begin{eqnarray}
    \text{Ad}_{\bm{T}} &=& \left[ \begin{array}{cc} 1 & \bm{0} \\ {[\bm{t}]} & \bm{R} \end{array} \right] \in \Re^{3 \times 3}, \label{eqn:Ad} \\
    \text{Ad}_{\bm{T}}^* &=& \left[ \begin{array}{cc} 1 & -{[\bm{t}]}^T \bm{R} \\ \bm{0} & \bm{R} \end{array} \right] \in \Re^{3 \times 3}, \label{eqn:dAd} \\
    \text{ad}_{\bm{\xi}}^* &=& \left[ \begin{array}{cc} 0 & -{[\bm{v}]}^T \\ \bm{0} & [w] \end{array} \right] \in \Re^{3 \times 3}. \label{eqn:dad}
\end{eqnarray}
where $[\bm{t}]$ is defined for $\bm{t} = (t_x, t_y) \in \Re^2$ as
\begin{equation}
    [\bm{t}] = \left[ \begin{array}{c} t_y \\ -t_x \end{array} \right] \in \Re^2.
    \label{eqn:skew2}
\end{equation}
Note that the notation $[\cdot]$ is defined differently for a scalar and $2$D vector as in (\ref{eqn:skew1}) and (\ref{eqn:skew2}). 
Both definitions are $2$D reductions of the $3 \times 3$ skew-symmetric matrix representation of $3$D rotation and translation vectors, respectively. Consequently, it inherits the anticommutative property of the cross product as $[w] \bm{t} = -[\bm{t}] w$ for $w \in \Re$ and $\bm{t} \in \Re^2$.

%========================================================================%
\subsection{Explicit Matrix Formula of \texorpdfstring{$\bm{A}_i, \bm{B}_i, \bm{C}_i, \bm{D}_i$}{A\_i, B\_i, C\_i, D\_i} and \texorpdfstring{$\bm{E}_i$}{E\_i}}

The matrices $\bm{A}_i, \bm{B}_i, \bm{C}_i, \bm{D}_i$ and $\bm{E}_i$ are given as
\begin{equation}
\begin{split}
    \bm{A}_i &= -\text{Ad}_{\bm{T}_{i}^{-1} \bm{T}_{i-1}} \\
    \bm{B}_i &= -\left[ \begin{array}{cc}
                \text{Ad}_{\bm{T}_{\text{p},i}} 
                    (\bm{\xi}_{\text{c},i-1} - \bm{\xi}_{\text{p},i})
                    & \bm{0}_{3 \times 2}   
            \end{array} \right] \\
    \bm{C}_i &= \frac{\partial \bm{\mathcal{F}}_{\text{ext},i}}{\partial \bm{\xi}_i} \\
    \bm{D}_i &= \left[ \begin{array}{cc} \bm{D}_{s,i} & \bm{D}_{\bm{f},i} \end{array} \right], \ 
    \bm{E}_i = \left[ \begin{array}{cc} \bm{E}_{s,i} & \bm{E}_{\bm{f},i} \end{array} \right]
\end{split}
\label{eqn:ABCDE}
\end{equation}
where
\begin{equation}
\begin{split}
    &{
    \bm{D}_{s,i} = \sum_{j \in \{ \text{l}, \text{r} \}} 
    \text{Ad}_{\bm{\mathcal{I}}(\bm{c}_{j,i})}^* \left[ \begin{array}{c} 0 \\ \tau_j \frac{\partial \hat{\bm{v}}_{j,i}}{\partial s_i} \end{array} \right]
    - \text{Ad}_{\bm{T}_{\text{c},i}}^* \text{ad}_{\bm{\xi}_{\text{c},i}}^* \left[ \begin{array}{c} 0 \\ \bm{f}_{i} \end{array} \right] } \\
    &\bm{D}_{\bm{f},i} = -\text{Ad}_{\bm{T}_{\text{c},i}}^* \left[ \begin{array}{c} \bm{0}_{1 \times 2} \\ \bm{I}_{2 \times 2} \end{array} \right] \\
% \end{split}
% \end{equation}
% and
% \begin{align}
% \begin{split}
    &{
    {\bm{E}}_{s,i} = \sum_{j \in \{ \text{l}, \text{r} \}} 
    \text{Ad}_{\bm{\mathcal{I}}(\bm{p}_{j,i})}^* \left[ \begin{array}{c} 0 \\ \tau_j \frac{\partial \hat{\bm{w}}_{j,i}}{\partial s_{i-1}} \end{array} \right]
    + \text{Ad}_{{\bm{T}}_{\text{p},i}}^* \text{ad}_{\bm{\xi}_{\text{p},i}}^* \left[ \begin{array}{c} 0 \\ {\bm{f}}_{i-1} \end{array} \right] } \\
    &\bm{E}_{\bm{f},i} = \text{Ad}_{\bm{T}_{\text{p},i}}^* \left[ \begin{array}{c} \bm{0}_{1 \times 2} \\ \bm{I}_{2 \times 2} \end{array} \right].
\end{split}
\end{equation}
Exceptionally, $\bm{D}_n$ is defined as $\bm{D}_n = \bm{I}$. 

%========================================================================%
\subsection{Differentiation of External Loads}
\label{app:loads}

We assume that the external load $\bm{\mathcal{F}}_{\text{ext},i}$ is a function of the link frame $\bm{T}_i$. The differentiation of the load, $\frac{\partial \bm{\mathcal{F}}_{\text{ext},i}}{\partial \bm{\xi}_i}$ is involved in (\ref{eqn:dequilibrium}) as well as the computation of $\bm{C}_i$ in (\ref{eqn:ABCDE}). Although we could not cover all possible load functions, a set of load examples will be discussed for readers to refer to in their own force implementations.

For describing a general differentiation procedure, we consider $\bm{\mathcal{F}}_{\text{ext},i}$ as a function of $\bm{T}_i$:
% Before providing examples, let us describe a general differentiation procedure.
% Consider $\bm{\mathcal{F}}_{\text{ext},i}$ as a function of $\bm{T}_i$: 
\begin{equation}
    \bm{\mathcal{F}}_{\text{ext},i} = \bm{\mathcal{F}}_{\text{ext},i}(\bm{T}_i).
\end{equation}
The differentiation of a function is derived as the first-order coefficient in the Taylor expansion. One can substitute $\bm{T}_i + \bm{T}_i [\delta \bm{\xi}_i]$ into $\bm{T}_i$ in the above function and expand it up to the first order. Subsequently, the first-order coefficient is taken as the differentiation.

%------------------------------------------------------------------------%
\subsubsection{Constant Body-Frame Load}

Although, in the real world, nonzero constant body-frame loads barely exist, we discuss this load type because it covers the unload case (i.e., constantly zero-force case). Obviously, the differentiation of a constant function is zero:
\begin{equation}
    \frac{\partial \bm{\mathcal{F}}_{\text{ext},i}}{\partial \bm{\xi}_i} = 0.
\end{equation}
% This case decouples the second row of (\ref{eqn:diff}) from the first row. More precisely, $\bm{C}_i$ becomes $\bm{C}_i = 0$; thus, the second row becomes an equation of only $\delta \bm{\eta}_i$ and $\delta \bm{\eta}_{i-1}$.
%
% This case decouples the second row of (\ref{eqn:diff}) from the first row because $\bm{C}_i$ becomes $\bm{C}_i = \bm{0}$. One can further optimize the iterative algorithm in this case, whereas it will not be discussed in detail for minimal distraction.
%
This case decouples the second row of (\ref{eqn:diff}) from the first row because $\bm{C}_i$ becomes $\bm{C}_i = \bm{0}$. This allows for further optimization of the iterative algorithm, whereas this optimization is not discussed in detail to avoid unnecessary distraction.

%------------------------------------------------------------------------%
\subsubsection{Constant Workspace Load}

We discuss this load type because it covers gravity forces. Let $\bm{\mathcal{F}}_\text{ws} = (m_\text{ws}, \bm{f}_{\text{ws}}) \in \Re^3$ denote the constant load expressed in the global frame. Suppose that the load is exerted at a body-fixed point of the link, denoted by $\bm{q} \in \Re^2$. Using (\ref{eqn:dAdf}), the external load is expressed in the link's body frame as
\begin{equation}
    \bm{\mathcal{F}}_{\text{ext},i}(\bm{T}_i) = \text{Ad}^*_{\bm{X}} \bm{\mathcal{F}}_\text{ws},
    \label{eqn:Fws}
\end{equation}
where
\begin{equation}
    \bm{X} = \left[ \begin{array}{cc}
                    \bm{R}_i^T & \bm{q} \\
                    \bm{0} & 1
                    \end{array} \right].
\end{equation}
Considering a small perturbation $\delta \bm{\xi}_i = (\delta w_i, \delta \bm{v}_i)$ in $\bm{T}_i$ and using the differentiation in (\ref{eqn:diff_dAd}), Equation (\ref{eqn:Fws}) reduces to
\begin{equation}
    \delta \bm{\mathcal{F}}_{\text{ext},i} = 
    \text{Ad}_{\bm{X}}^* \text{ad}_{\delta \bm{\xi}_{\bm{X}}}^* \bm{\mathcal{F}}_\text{ws}
\end{equation}
where $\delta \bm{\xi}_{\bm{X}} = (-\delta w_i, \bm{0})$. Substituting (\ref{eqn:dAd}) and (\ref{eqn:dad}) into the above equation and utilizing the anticommutative property $[\delta w_i] \bm{f}_{\text{ws}} = -[ \bm{f}_{\text{ws}} ] \delta w_i$, we can obtain
\begin{equation}
    \delta \bm{\mathcal{F}}_{\text{ext},i} = 
    \left[ \begin{array}{c}
    -{[\bm{q}]}^T \bm{R}_i^T [ \bm{f}_{\text{ws}} ] \\ \bm{R}_i^T [ \bm{f}_{\text{ws}} ]
    \end{array} \right] \delta w_i.
\end{equation}
Finally, the differentiation is given as
\begin{equation}
    \frac{\partial \bm{\mathcal{F}}_{\text{ext},i}}{\partial \bm{\xi}_i} = \left[ 
    \begin{array}{ccc}
    \begin{array}{c}
    -{[\bm{q}]}^T \bm{R}_i^T [ \bm{f}_{\text{ws}} ] \\ \bm{R}_i^T[ \bm{f}_{\text{ws}} ]
    \end{array}
    & \bm{0} & \bm{0}
    \end{array}
    \right].
\end{equation}

%------------------------------------------------------------------------%
\subsubsection{Linear Elastic Force}

A spring force is discussed here to demonstrate how a non-constant load can be differentiated.
If link $i$ is connected with a fixed workspace point $\bm{q} \in \Re^2$ through a linear spring, the external load is given as
\begin{equation}
    \bm{\mathcal{F}}_{\text{ext},i}(\bm{T}_i) = \left[ \begin{array}{c}
    0 \\ -k \bm{R}_i^T (\bm{t}_i - \bm{q})
    \end{array} \right],
\end{equation}
where $k \in \Re$ is the spring coefficient. Considering a small perturbation $\delta \bm{\xi}_i = (\delta w_i, \delta \bm{v}_i)$, the above load expands as
\begin{equation}
    \delta \bm{\mathcal{F}}_{\text{ext},i} = \left[ \begin{array}{c}
    0 \\ -k \bm{R}_i^T [-\delta w_i] (\bm{t}_i - \bm{q}) - k \delta \bm{v}_i
    \end{array} \right].
\end{equation}
As it holds $[-\delta w_i] (\bm{t}_i - \bm{q}) = [\bm{t}_{i} - \bm{q}] \delta w_i$, the differentiation is derived as
\begin{equation}
    \frac{\partial \bm{\mathcal{F}}_{\text{ext},i}}{\partial \bm{\xi}_i} = \left[ 
    \begin{array}{cc}
    0 & \bm{0} \\
    - k \bm{R}_i^T [\bm{t}_{i} - \bm{q}] & -k \bm{I}
    \end{array}
    \right].
\end{equation}

%%%%%%%%%%%%%%%%%%%%%%%%%%%%%%%%%%%%%%%%%%%%%%%%%%%%%%%%%%%%%%%%%%%%%%%%%%%%%%%%

\bibliography{references}
\bibliographystyle{ieeetr}

%======================================================================%

% \clearpage

\end{document}